\documentclass[10pt,twocolumn,letterpaper]{article}

\usepackage{iccv}
\usepackage{times}
\usepackage{epsfig}
\usepackage{graphicx}
\usepackage{amsmath}
\usepackage{amssymb}
\usepackage{placeins}
\usepackage[bottom]{footmisc} 
\usepackage[accsupp]{axessibility}
\usepackage{float}

\usepackage[pagebackref=true,breaklinks=true,letterpaper=true,colorlinks,bookmarks=false]{hyperref}

\usepackage[ruled,vlined,linesnumbered]{algorithm2e}
\SetKwInput{KwInput}{Input}
\SetKwInput{KwOutput}{Output}

\SetCommentSty{mycommfont}

\iccvfinalcopy 

\def\iccvPaperID{10412} 

\ificcvfinal\fi

\begin{document}

\title{SemIE: Semantically-aware Image Extrapolation}

\author{
    Bholeshwar Khurana$^{1,2}$\thanks{BK, SRD, AB contributed to the work during their Adobe internship.} \and
    Soumya Ranjan Dash$^{1,2*}$\and
    Abhishek Bhatia$^{1,2*}$ \and
    \vspace{0.1cm}
    Aniruddha Mahapatra$^2$\and
    Hrituraj Singh$^{2,3}$\and
    Kuldeep Kulkarni$^2$\and
    $^1$IIT Kanpur \quad $^2$Adobe Research India \quad $^3$Triomics
\\{
    \tt\small \{bkhurana, sodash, abbhatia, anmahapa, kulkulka\}@adobe.com,
    hrituraj@triomics.in
}
}

\maketitle
\ificcvfinal\thispagestyle{empty}\fi

\begin{abstract}
   We propose a semantically-aware novel paradigm to perform image extrapolation that enables the addition of new object instances. All previous methods are limited in their capability of extrapolation to merely extending the already existing objects in the image. However, our proposed approach focuses not only on \textbf{(i)} extending the already present objects but also on \textbf{(ii)} adding new objects in the extended region based on the context. To this end, for a given image, we first obtain an object segmentation map using a state-of-the-art semantic segmentation method. The, thus, obtained segmentation map is fed into a network to compute the extrapolated semantic segmentation and the corresponding panoptic segmentation maps. The input image and the obtained segmentation maps are further utilized to generate the final extrapolated image. We conduct experiments on Cityscapes and ADE20K-bedroom datasets and show that our method outperforms all baselines in terms of FID, and similarity in object co-occurrence statistics. Project url: \url{https://semie-iccv.github.io/}
\end{abstract}
\begin{figure}[t]
\begin{center}
\includegraphics[width=0.9\linewidth]{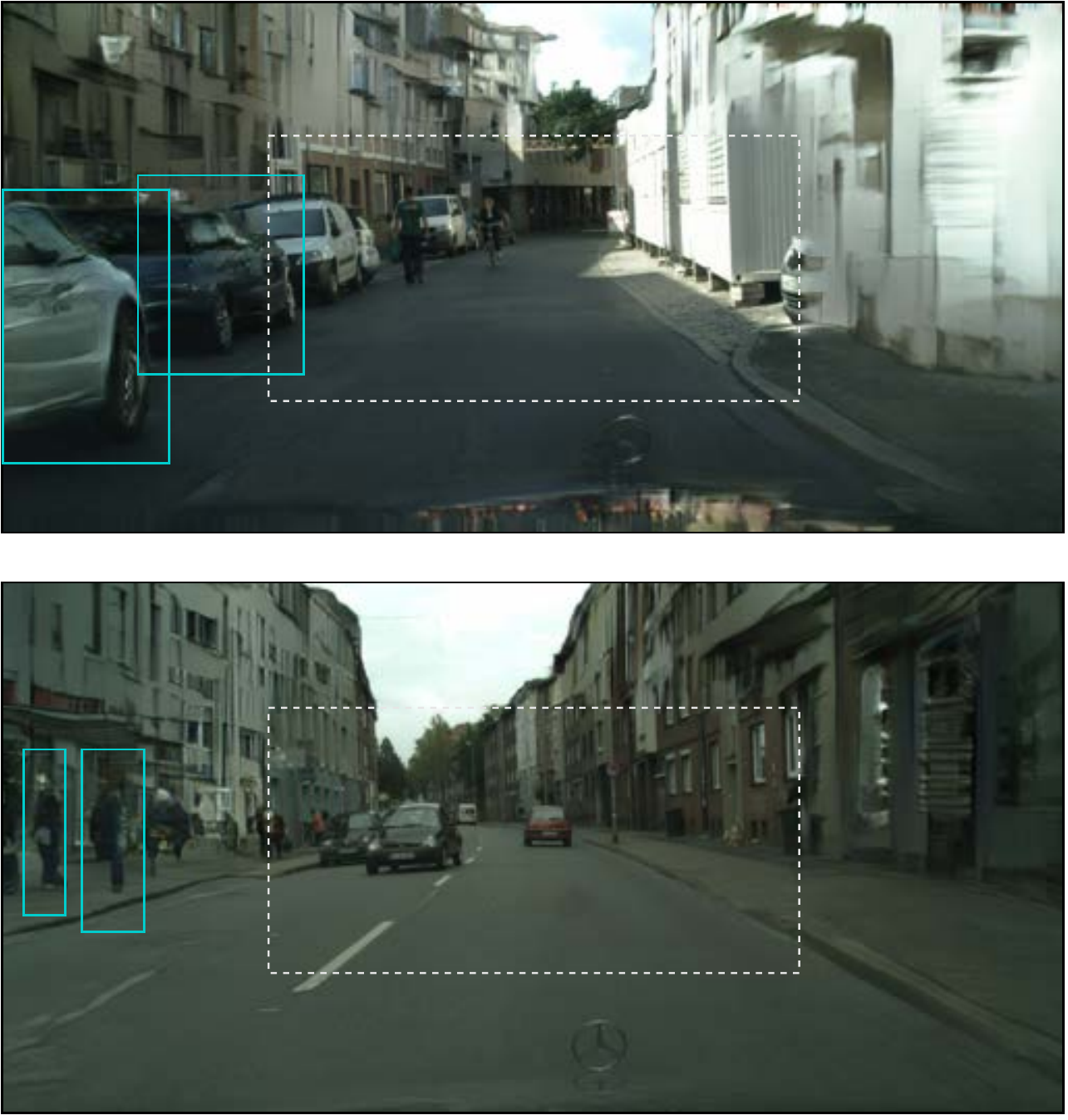}
\end{center}
    \vspace{-5pt}
   \caption{Illustration of our results. Dotted white rectangle refers to the input image. Our method not only extrapolates the objects present in the input but also generates new objects (blue bounding boxes) while maintaining the texture consistency.}
\vspace{-15pt}
\end{figure}
\vspace{-2pt}
\section{Introduction}
Image extrapolation or out-painting refers to the problem of extending an input image beyond its boundaries. While the problem has applications in virtual reality, sharing photos on social media like Instagram, and even generating scenes during game development especially if the scenes are repetitive, it is relatively under-explored compared to the image inpainting counterpart, which has been extensively researched. Image inpainting solutions based on deep networks and generative adversarial networks (GANs), when applied to the out-painting problem, have been shown to yield poor results \cite{teterwak2019boundless}. This has led to researchers exploring and proposing new solutions to the out-painting problem \cite{Zhang_2020, yang2019very, wang2019wide}. However, the solutions have been mainly restricted to images that involve outdoor domains like natural scenes where the problem is limited to just extending the existing textures for `stuff' classes like mountains, water, trees \cite{guo2020spiral, teterwak2019boundless} or single-object images of classes like faces, flowers, and cars. These methods are not suitable to other domains like traffic scenes and indoor scenes where a desirable image extrapolation necessitates 1) extending not only the `stuff' classes but also the `things' classes like cars, persons, beds, tables that have very definite structure as well as 2) adding new objects based on the context that were not present in the original image.  
So, why cannot we use the existing techniques \cite{Zhang_2020, yang2019very, wang2019wide, teterwak2019boundless} for such domains? The answer is they fail spectacularly by filling the extrapolated region with artifacts (see figures \ref{fig:baselines_city} and \ref{fig:baselines_ade}). They attempt to extrapolate the image by capturing the low-level statistics like textures and colors from the input image while ignoring the high-level information like object semantics and object co-occurrence relationships. In short, they are limited in their ability to perform satisfactory image extrapolation that demands the creation of new object instances and the extension of multiple objects from diverse classes.

We address the shortcomings of the previous works by extrapolating the image in the semantic label map space, which enables us to generate new objects in the extrapolated region. Additionally, semantic label maps belong to a lower dimensional manifold than images, making it easier to extrapolate them. However, just having a  semantic label map does not allow us to have control over every instance in the extrapolated image. We propose to generate an estimate of the panoptic label directly from the extrapolated semantic label map, different from \cite{kirillov2019panoptic, cheng2020panoptic}. Instance boundary maps obtained from panoptic labels also help in creating crisper boundaries between objects belonging to the same semantic category. Unlike semantic label map to image generation \cite{park2019semantic, isola2017image, wang2018high}, we have to maintain texture consistency between the input and the extrapolated regions. To account for this, we propose Instance-aware context normalization (IaCN), which leverages the estimated panoptic label maps to transfer instance-wise average color information as a feature map for texture consistency in the extrapolated parts of the corresponding object instances. In addition, we propose the use of patch co-occurrence discriminator \cite{park2020swapping} to maintain global texture similarity in input and extrapolated region. 

Our contributions can be summarized below: 
\vspace{-5pt}
\begin{itemize}
    \item We propose a novel paradigm for image out-painting by extrapolating the image in the semantic label space to generate novel objects in the extrapolated region.
    \vspace{-5pt}
    \item We propose the estimation of panoptic label maps from the extrapolated semantic label maps to facilitate the generation of high quality object boundaries in the extrapolated image.
    \vspace{-5pt}
    \item We propose Instance-aware Context Normalization (IaCN) and the use of patch co-occurrence discriminator to maintain texture consistency of extrapolated instances.
\end{itemize}
\vspace{-5pt}
Through extensive experiments on Cityscapes and ADE20K datasets, we show that our method outperforms all previous state-of-the-art methods in terms of FID and similarity in object co-occurrence metrics.

\begin{figure*}
\begin{center}
\includegraphics[width=0.9\linewidth]{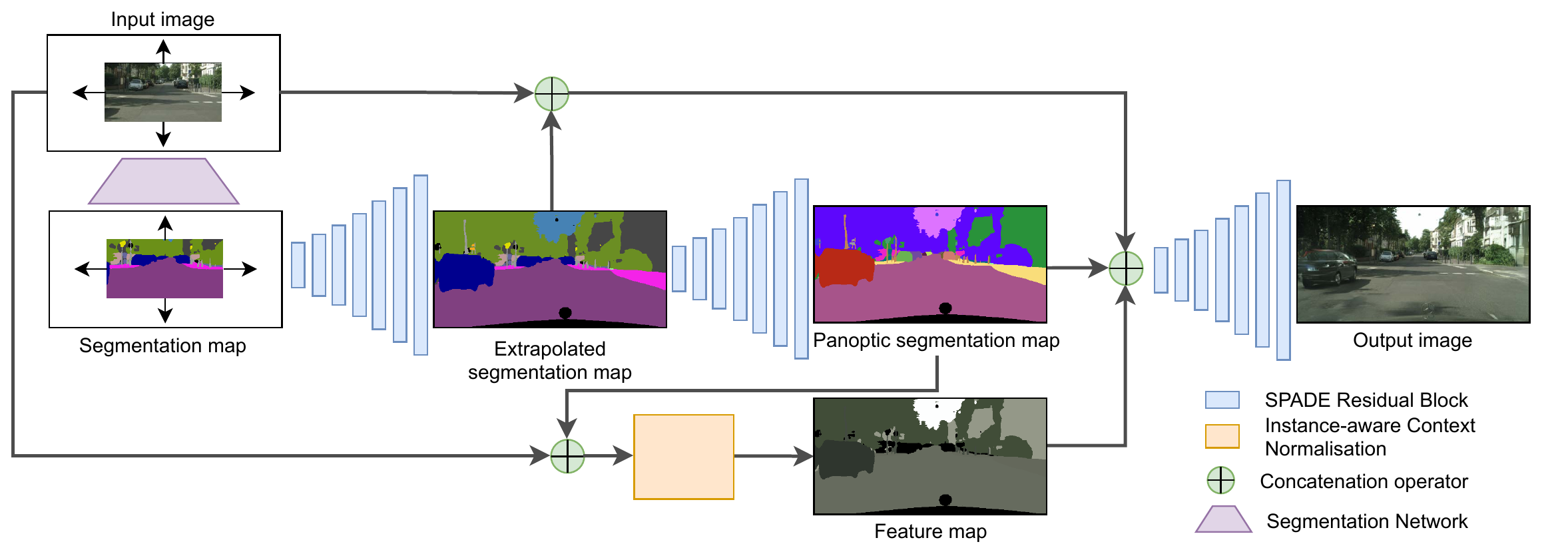}
\end{center}
    \vspace{-5pt}
   \caption{\textbf{Overview of the pipeline:} Stage 1: The input image is fed into a pre-trained segmentation network to obtain its label map. Stage 2: The stage 1 output fed into a network to obtain the extrapolated label map. Stage 3: The extrapolated label map is fed into another network to obtain the panoptic label map. Stage 4: The input image, extrapolated label map and the panoptic label map are used in conjunction with Instance-aware context normalization module to obtain the final extrapolated image.}
\label{fig:pipeline}
\vspace{-10pt}
\end{figure*}

\section{Related Work}
\noindent \textbf{Image Extrapolation:} Prior works in image synthesis have had great breakthroughs in image inpainting \cite{liu2018image, Yeh_2017_CVPR, yu2018generative} , conditional image synthesis \cite{isola2017image, liu2019learning, park2019semantic, wang2018high, wang2020sketchguided}, and unconditional image synthesis \cite{arjovsky2017wasserstein, mescheder2018training, durugkar2016generative}. On the contrary, image extrapolation models have been relatively less successful. The works on image extrapolation can be broadly classified based on whether they use non-parametric methods or parametric methods. Several non-parametric methods \cite{efros2001image, 790383} have been able to perform only a limited peripheral texture extension. Furthermore, their heuristics do not capture the variation in color, texture and the information of shape and structure of an object. These methods \cite{790383} limit themselves to simple pattern extrapolation and are very brittle to increasing extrapolation. Other classical approaches \cite{barnes2009patchmatch} leverage patch matching to extrapolate the image. However, these methods are limited in their ability to generate new objects or hallucinate new textures.
With the advent of GAN \cite{goodfellow2014generative} based approaches, significant progress has been made in image extrapolation. \cite{Zhang_2020, yang2019very, teterwak2019boundless} use a single-stage method to extrapolate the input image. Most of these works deal with scene completion using object completion or merely extending the significant texture near the image boundary. \cite{wang2019wide} proposed a method of feature expansion from input region to predict context for the extrapolated image, while \cite{guo2020spiralnet} generates the extrapolated image by incrementally extending image on each side using a generated reference image. Consequently, as we go further from the input boundary, the relative volume information from input reduces resulting in the generation of substandard extrapolation.
Moreover, all of these approaches currently lack the semantic understanding of the scene and semantic structure of objects in the scene. 
\\
\noindent \textbf{Semantic Editing for Image Manipulation:}
Recently, there have been a few works that manipulate images by editing in the semantic label space. \cite{azadi2019semantic} concentrates on synthesizing images using semantic label space but in an unconditional setting, where the semantic label space is generated from scratch from a random seed. \cite{hong2018learning, lee2018context} proposed methods to insert an object in an image by editing the semantic label of the input image, given the class information and the bounding box of the object. Such methods are unsuitable to be adapted for image out-painting as we do not have the class information and the bounding boxes for the new objects to be inserted in the extrapolated region. Another drawback of such methods is that they require as many forward passes as the number of the new objects and hence are not scalable in terms of time complexity. \cite{ntavelis2020sesame} proposes to edit an image by allowing the user to provide a semantic guideline of the regions to be manipulated. Different from the above works, our image extrapolation method automatically extrapolates the semantic label map of the image without any user input and estimates the corresponding panoptic label map for the extrapolated image to be generated. The closest work to ours is an in-painting model, SPGNet \cite{song2018spg} where the hole in an image is firstly filled in the semantic label space before the final image is generated. However, unlike SPGNet, we generate an estimate of panoptic label map, along with the semantic extrapolation, and further leverage it to ensure texture transfer for the extrapolated instances and sharp instance boundaries in the final image. We show through experiments that our method is significantly better than SPGNet.

\section{Our Method}
Our goal is to extrapolate a given image $ \mathbf{X} \in \mathbb{R}^{h\times w\times c}$ on its periphery using a sequence of deep neural networks. $\mathbf{Y} \in \mathbb{R}^{h_1\times w_1\times c}$ is the extrapolated image where $h_1\geq h$ and $w_1\geq w$. Here, $c$ represents the number of channels corresponding to the image, which is three for an RGB image. The pipeline shown in figure \ref{fig:pipeline} involves four major stages:
\vspace{-5pt}
\begin{itemize}
    \item Image segmentation: Generation of semantic label map from the input image.
    \vspace{-5pt}
    \item Semantic label map extrapolation: Extend periphery in the semantic label space.
    \vspace{-5pt}
    \item Panoptic label estimation: The semantic label map is processed to obtain an apriori estimate of corresponding panoptic label map. 
    \vspace{-5pt}
    \item Instance-aware image synthesis: Generation of image from the semantic label map and panoptic label map by leveraging the proposed IaCN module and patch co-occurrence discriminator.
\end{itemize}

\subsection{Image Segmentation}
Given an image $\mathbf{X}\in\mathbb{R}^{h\times w\times c}$, corresponding one-hot vector for semantic label map $\mathbf{L_{1}}\in\{0,1\}^{h\times w\times c_1}$ can be obtained using state-of-the-art segmentation techniques \cite{zhao2017pyramid, tao2020hierarchical, cheng2020panoptic, zhang2020resnest, yu2020context}. For our method, we use PSPNet \cite{zhao2017pyramid}. 

\begin{figure*}
\begin{center}
\includegraphics[width=\linewidth]{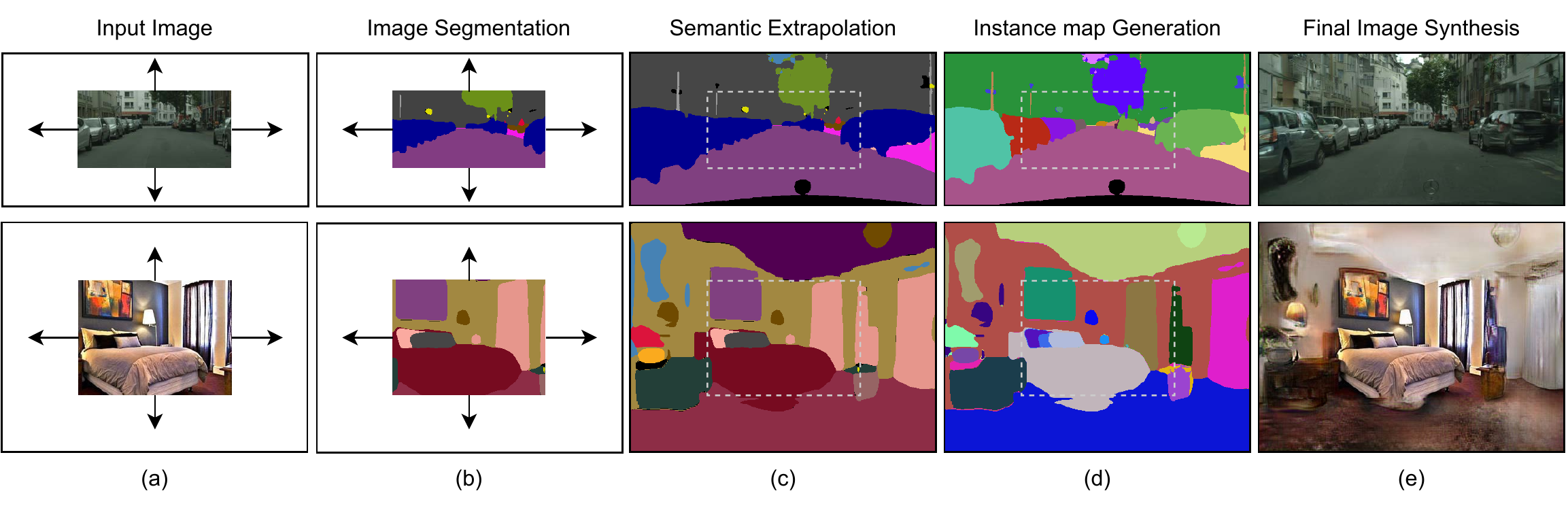}
\end{center}
    \vspace{-5pt}
   \caption{\textbf{Stage-wise results:} The input (cropped) image is converted to semantic label map in stage 1, which is then extrapolated in stage 2 to form the outpainted semantic label map. Panoptic label maps are generated from this semantic label map in stage 3. The input image, the (outpainted) semantic label map and the panoptic label map are used to synthesize the final image in stage 4.}
\label{fig:stagewise_result}
\vspace{-10pt}
\end{figure*}

\subsection{Semantic Label Extrapolation}
\label{semantic}
We train a network, dubbed `Peripheral Object Generation (POGNet)', $G_{\mathbf{S}}$ to semantically extrapolate $\mathbf{L_1}$ and obtain an estimate of the semantic label map, $\mathbf{L_2}$ of the final extrapolated image to be generated. In addition to generating $\mathbf{L_2}$, we also output the corresponding instance boundary channel. Although \cite{song2018spg} uses input image with semantic label map to generate extrapolated semantic label map, having explicit supervision with ground truth instance boundary map acts as a better regularizer during training for obtaining more precise object shapes. POGNet is trained using a multi-scale discriminator as proposed in \cite{wang2018high}, enabling $G_{\mathbf{S}}$ to capture the object co-occurrence information at various scales. 
\\
\textbf{Adversarial Loss:}
Instead of regular GAN loss \cite{goodfellow2014generative}, we use LS-GAN loss \cite{mao2017least} ($\mathbf{\mathcal{L}}_{GAN}$).

\noindent \textbf{Focal Loss:} We use focal loss to compute the discrepancy between the ground truth semantic label map and the output of the POGNet. By giving higher weight to hard-to-generate object classes, focal loss allows us to generate some of the rare classes. The focal loss between the ground-truth and the output at any location is given as:
\vspace{-5pt}
\[ l(z, y) = -y \times log(z) \] 
\[  \mathcal{L}_{CE}(z, y) = \Sigma_{h,w,c} l(z,y)\]
\[  \mathcal{L}_{FL}(z, y) = l(z,y) \times (1-z)^\gamma\]
The final focal loss, $\mathcal{L}_{FL}^{all}$ is given by the sum of focal losses across all locations in the semantic label map. 
We use the following training objective for semantic label map extrapolation (we show only the generator losses here):
\vspace{-5pt}
\begin{equation}
\label{eqn:stage2_obj}
\begin{split}
     \mathcal{L}_{gen} = \mathcal{L}_{GAN} + \mathcal{L}_{FM} + \lambda_{FL}\mathcal{L}_{FL}^{all} + \lambda_{CE}\mathcal{L}_{CE},
\end{split}
\vspace{-5pt}
\end{equation}
where $\mathcal{L}_{CE}$ is the cross-entropy loss between the ground-truth instance boundary and the corresponding output channel in POGNet and $\mathcal{L}_{FM}$ is the discriminator feature matching loss. More details can be found in Section \ref{sec:stage2_obj}.

\subsection{Panoptic Label Map Generation}
As mentioned earlier, we wish to estimate the panoptic label maps (for the to-be-generated extrapolated image) that can be leveraged for IaCN module (discussed in \ref{sec:iacn}) as well as obtain crisp and precise boundaries between the object instances. Traditionally, the panoptic label maps are generated from the images. But how do we estimate panoptic label maps, apriori, without knowing the image itself? We adapt the method elucidated in Panoptic-DeepLab \cite{cheng2020panoptic} by predicting the class-agnostic pixel-wise instance center maps and off-set maps from the instance centers for objects belonging to `things' classes, directly from the semantically extrapolated map, i.e the output of POGNet. Specifically, we train a generator-only network that takes in the extrapolated segmentation map and produces heat maps for instance centers and the pixel-wise offsets from the nearest instance center. The center heat-maps and the offset outputs are further processed along with the segmentation map to obtain the instance maps. The ground-truth center maps are represented by Gaussian blobs of standard deviation of 8 pixels, centered at the instance centers. We use $L_2$ loss to compute the instance center loss and $L_1$ loss to compute the offset losses. The final loss for stage 3 is the weighted sum of the center loss and the offset losses. 

During the test time, we adapt the procedure mentioned by \cite{cheng2020panoptic} to group the pixels based on the predicted centers and off-sets to form instance masks. The instance masks and the semantic label map (the input to stage 3) are combined by majority voting to obtain the panoptic label map. An expanded version of the details of training of the network and post-processing are provided in Section \ref{sec:stage3_supp}.

\subsection{Instance-aware Image Synthesis}
This is the final stage (stage 4) which converts the extrapolated semantic label map back into a colored image. This stage takes in input $\mathbf{X'} (\in \mathbb{R}^{h_1 \times w_1 \times c'}) $ (Figure \ref{fig:pipeline}), which is concatenation of the extrapolated semantic label map obtained from the second stage, the cropped (input) image, the boundary map obtained using the panoptic label map obtained from the previous stage and the feature map obtained using the proposed Instance-aware Context Normalization. The output is an RGB image $\mathbf{Y} \in \mathbb{R}^{h_1 \times w_1 \times 3}$.

This is different from prior conditional GANs problems \cite{isola2017image, liu2019learning, park2019semantic, wang2018high} since they synthesize RGB images from semantic label maps, but we have to synthesize RGB images from semantic label maps, given some pixel information of the to-be-synthesized RGB image, which is the cropped image in our case. Here, we have to take care of texture consistency in the synthesized image while maintaining an identity mapping from the input image to the corresponding part in the final image.

\noindent \textbf{Generator}\newline
We use SPADE \cite{park2019semantic} normalization residual blocks for each of the layers in the generator. We use similar learning objective functions, as used in SPADE \cite{park2019semantic} and pix2pixHD \cite{wang2018high}: GAN loss with hinge-term \cite{lim2017geometric, miyato2018spectral, zhang2019selfattention} ($\mathcal{L}_{GAN}$), Feature matching loss \cite{wang2018high} based on the discriminator ($\mathcal{L}_{FM}$) and VGGNet \cite{simonyan2015deep} for perceptual losses \cite{DB16c, johnson2016perceptual} ($\mathcal{L}_{VGG}$)

\noindent \textbf{Instance-aware Context Normalization (IaCN)}\newline
\label{sec:iacn}Outpainting-SRN \cite{wang2019wide} proposed Context Normalization (CN) to maintain texture consistency between the inside (cropped) region and the outside (outpainted) region. It involves transferring the mean feature or color from the inside region to the outside region. However, we believe that transferring this input mean color directly to the outside region is not suitable for images that have very diverse object instances (like outdoor images, street images).

To this end, we propose Instance-aware Context Normalization (IaCN) (Figure \ref{fig:pipeline}), which takes as input the cropped image and the instance map. IaCN module computes the mean color using the input (cropped) image for all the partial instances. Partial instances refer to the instances which get extrapolated in the final image. Since the problem with texture consistency occurs only for partial instances, therefore we compute features only for them. These computed feature maps are then concatenated to the input.

\begin{figure}[t]
\begin{center}
\includegraphics[width=0.9\linewidth]{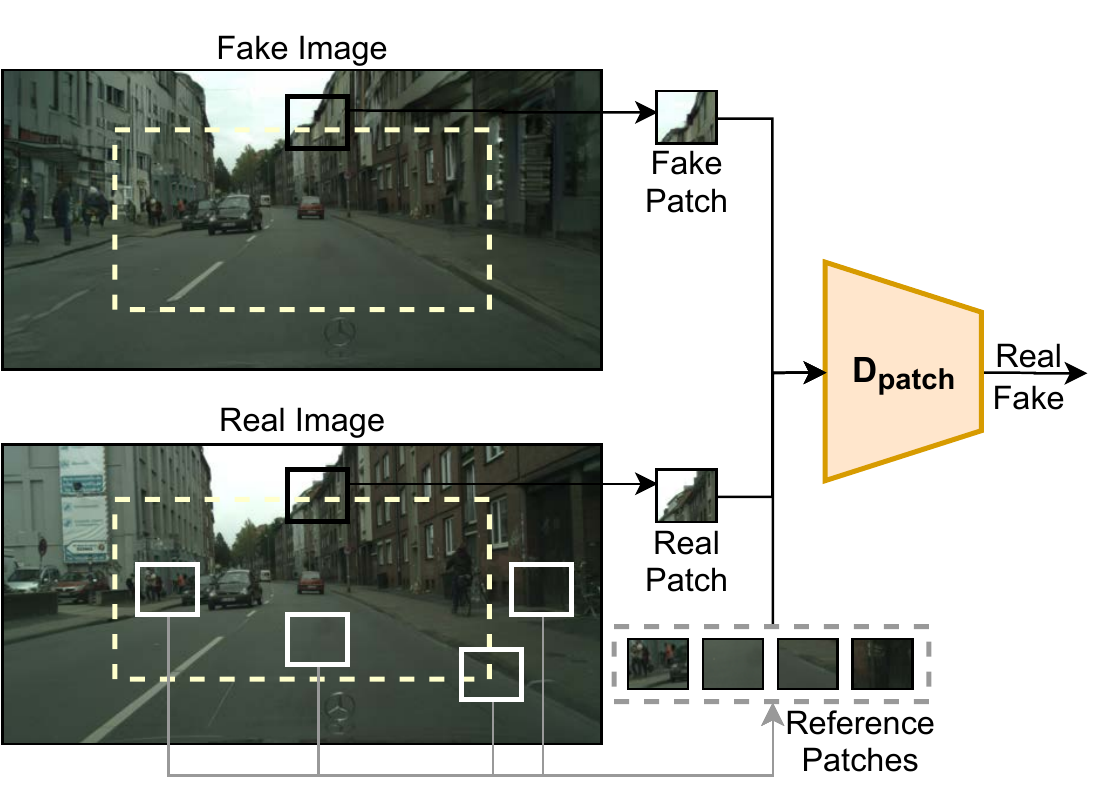}
\end{center}
    \vspace{-5pt}
   \caption{\textbf{Patch Discriminator}: $D_{patch}$ takes in input 4 reference patches, a fake patch and a real patch. The reference patches are randomly selected from the real image. The fake patch and real patch are the same patches, randomly selected but made sure that some part of them is inside while other part is outside, from fake image and real image respectively. The discriminator tries to distinguish between fake patch and the real patch, making use of the reference patches. All the patches are of size $64 \times 64$.}
\label{fig:D_patch}
\end{figure}

\begin{figure*}
\begin{center}
\includegraphics[width=\linewidth]{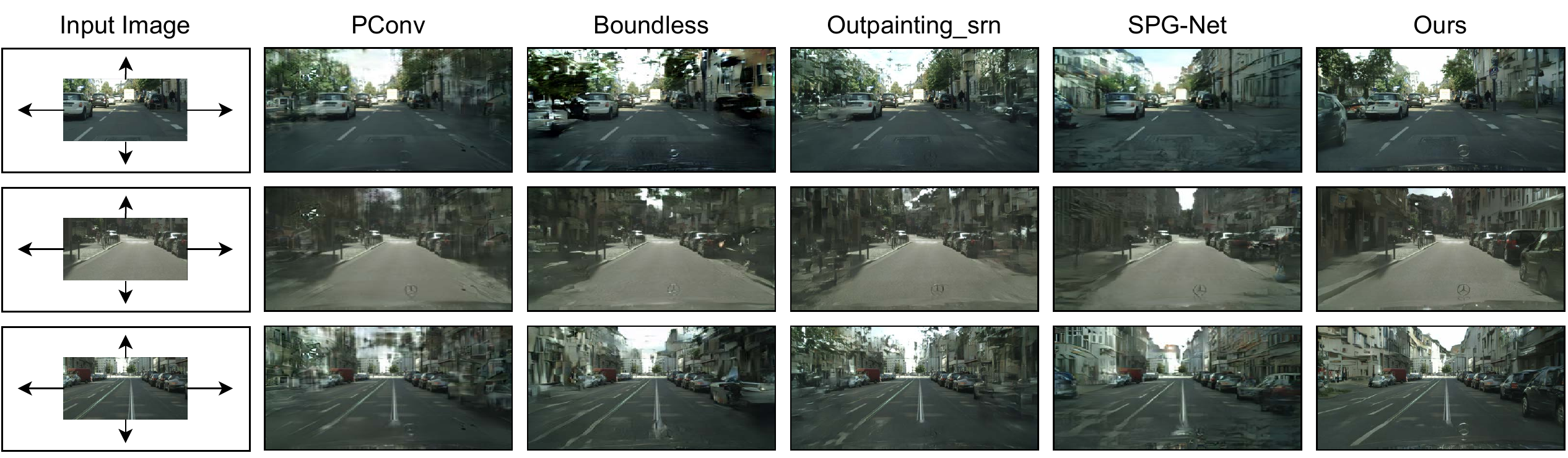}
\end{center}
    \vspace{-5pt}
    \caption{\textbf{Cityscapes dataset:} Our method is able to generate new objects in the extrapolated region leading to realistic image extrapolation. Except ours and SPGNet, all other methods fail to generate new objects in the extrapolated region.}
\label{fig:baselines_city}
\vspace{-10pt}
\end{figure*}

\begin{figure*}
\begin{center}
\includegraphics[width=\linewidth]{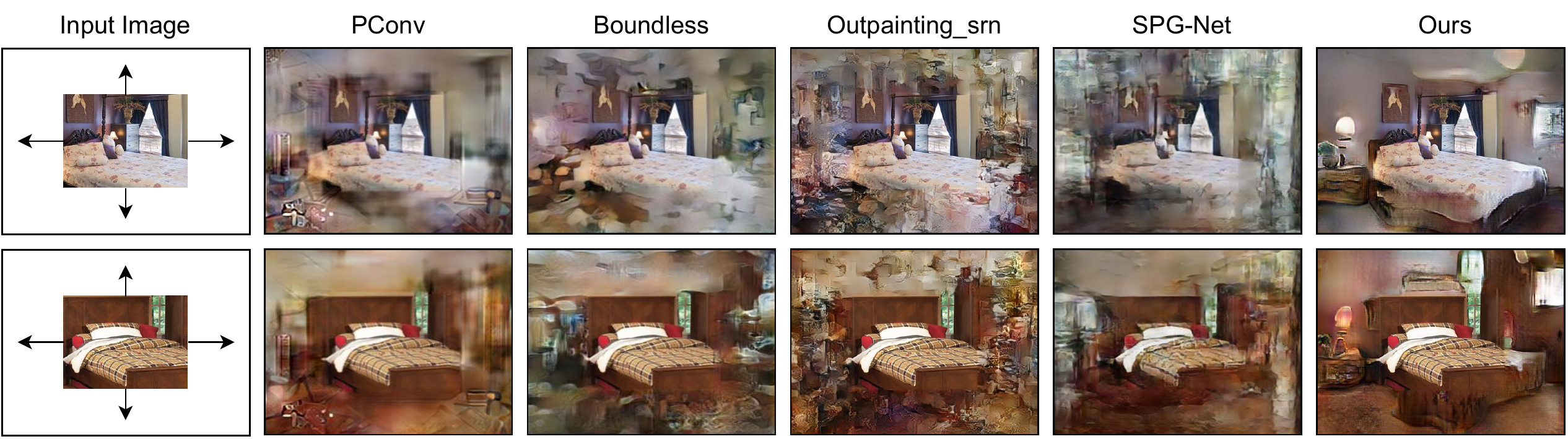}
\end{center}
    \vspace{-5pt}
  \caption{\textbf{ADE20K dataset:} Our method is able to generate new objects in the extrapolated region leading to realistic image extrapolation. Only our, all other methods try to copy texture patches from inside region in the extrapolated region.}
\label{fig:baselines_ade}
\vspace{-10pt}
\end{figure*}

\noindent \textbf{Discriminators} \newline
We propose to use two discriminators, i) a traditional image discriminator (multi-scale discriminator) that attempts to differentiate between the real and the fake image, ii) a patch co-occurrence discriminator similar to \cite{park2020swapping}, which employed it to ensure texture transfer \cite{karras2020analyzing, xian2018texturegan} from an input image to the target image to be edited. We employ a similar idea wherein the region in the image that needs to be extrapolated takes the role of the target image (equation \ref{eq:swap_1}). This facilitates consistent texture transfer from the inside region (source) to the extrapolated region (target) (illustrated in Figure \ref{fig:D_patch}).
\vspace{-5pt}
\begin{equation}
\label{eq:swap_1}
\begin{split}
    &\mathcal{L}_{CooccurGAN}\left(G, D_{patch}\right) = \\ &\mathbb{E}_{x,y}[-log(D_{patch}(crop(G(x)), crop(y), crops(y)))]
\end{split}
\end{equation}
Here $x$ is the input and $y$ is the corresponding ground-truth image. $crop(y)$ function takes a random patch from image $y$ and $crops(y)$ takes 4 random patches from image $y$, which serve as the reference patches.

The details of the network architectures for all generators and discriminators for the various stages are provided in Section \ref{sec:net_arch}.


\begin{table*}[t]
\begin{center}
\begin{tabular}{|l|c|c|c|c|c|c|}
\hline
Method & (Bed, Lamp) & (Wall, Window) & (Bed, Curtain) & (Floor, Table) & (Wall, Painting)\\
\hline\hline
Outpainting-SRN & 0.66 & 0.82 & \textbf{0.94} & 0.77 & 0.64\\
Boundless & 0.79 & 0.82 & 0.87 & 0.75 & 0.76\\
Pconv & 0.75 & 0.85 & 0.83 & 0.77 & 0.83\\
SPGNet & 0.77 & 0.53 & 0.51 & 0.84 & 0.82\\
SPGNet++ & 0.79 & 0.87 & 0.85 & 0.81 & 0.83\\
\textbf{Ours} & \textbf{0.82} & \textbf{0.90} & 0.84 & \textbf{0.87} & \textbf{0.84}\\
\hline
\end{tabular}
\end{center}
\vspace{-5pt}
\caption{\textbf{Results}: Similarity in object co-occurrence scores (higher is better) for our method vs the baselines on ADE20K-bedroom dataset.}
\label{table:socc_ade20k}
\end{table*}

\noindent \textbf{Variational Autoencoder} \newline
To ensure appropriate style transfer from the input image, we use an encoder that processes the input image, which is then fed to the generator. We use the encoder used in \cite{park2019semantic}. This encoder forms a VAE \cite{kingma2013auto} with the generator. In the objective function, we add a KL-Divergence Loss term \cite{kingma2013auto} ($\mathcal{L}_{KLD}$).
\\
\\
\\
\noindent \textbf{Final Objective} \newline
The training objective is as shown below in equation \ref{eqn:stage3_obj}:
\vspace{-5pt}
\begin{equation}
\label{eqn:stage3_obj}
\begin{split}
    \min\limits_{G} \{ &\mathcal{L}_{GAN} + \lambda_{FM}\mathcal{L}_{FM} + \lambda_{VGG}\mathcal{L}_{VGG} \\ & + \lambda_{KLD}\mathcal{L}_{KLD} + \mathcal{L}_{CooccurGAN} \}
\end{split}
\vspace{-5pt}
\end{equation}

\section{Experiments}
We evaluate the proposed approach on two different datasets which have a sufficient disparity between each other to show that our approach is fairly robust and is applicable to diverse scenes. We utilize the publicly available Cityscapes \cite{cordts2016cityscapes} and ADE20K-bedroom subset \cite{Zhou_2017_CVPR} datasets both of which contain large variety of distinct object categories. While Cityscapes comprises of outdoor street images, ADE20K bedroom subset consists of bedroom scenes. The ADE20K processed subset\footnote{To obtain the processed subset, we contacted the authors of \cite{kulkarni2020hallucinet}.} consists of 31 classes including bed, lamp, wall, floor and table. Cityscapes consists of 2975 training images and 500 validation images. Each image has its corresponding semantic label map and instance label map along with the original image. The bedroom subset of ADE20K \cite{Zhou_2017_CVPR} has 1389 images in the training set and 139 in the validation set. In order to limit the size of our model, we downsample the images in Cityscapes to a resolution of $ 256 \times 512$ and the ADE20K bedroom by resizing all its images to a standard size of $384 \times 512$ while training. For both the datasets, the input image is taken as centre crop of resized image with half the height and width.

\noindent \textbf{Implementation details} \newline
We train PSPNet \cite{zhao2017pyramid} on Cityscapes as well as ADE20K bedroom subset at the resolution discussed earlier and use them to generate segmentation maps of the input (cropped) images. We adopt cGAN based generator for stage 1, stage 2 and stage 4 models similar to SPADE \cite{park2019semantic}. In stage 2 we replace $tanh$ with $sigmoid$ activation in the final layer to produce one hot encodings and semantic label map. For the training of stage 2, in our final objective (Eq. \ref{eqn:stage2_obj}), we use $\lambda_{FL} = 5$, $\lambda_{CE} = 5 $ and $\gamma = 5$. For the training of stage4, we use the same weights for loss terms as \cite{park2019semantic}, i.e. $\lambda_{FM} = 10, \lambda_{VGG} = 10 \text{ and } \lambda_{KLD} = 0.05$ in Eq \ref{eqn:stage3_obj}. We use ADAM solver \cite{kingma2014adam} with $\beta_1 = 0$ and $\beta_2 = 0.9$ for both the stages. The training is done for 200 epochs.

\begin{table*}[t]
\begin{center}
\begin{tabular}{|l|c|c|c|c|c|}
\hline
Method & (Parking, Car) & (Person, Person) & (Pole, Traffic Light) & (Person, Rider) & (Car,Sidewalk)\\
\hline\hline
Outpainting-SRN & 0.85 & 0.89 & 0.68 & 0.91 & 0.85\\
Boundless & 0.82 & 0.89 & \textbf{0.99}  & 0.94 & 0.82\\
Pconv & 0.83 & 0.88 & 0.57 & 0.9 & 0.83\\
SPGNet & 0.91 & 0.84 & 0.87 & 0.86 & 0.91\\
SPGNet++ & 0.94 & 0.87 & 0.93 & 0.94 & 0.93\\
\textbf{Ours} & \textbf{0.96} & \textbf{0.92} & 0.96 & \textbf{0.96} & \textbf{0.94}\\
\hline
\end{tabular}
\end{center}
\vspace{-5pt}
\caption{\textbf{Results}: Similarity in object co-occurrence scores (higher is better) for our method vs the baselines on Cityscapes dataset.}
\label{table:socc_cityscapes}
\vspace{-10pt}
\end{table*}

\noindent \textbf{Baselines} \newline
We compare our method with various baselines both in quantitative (with FID and Similarity in Object Co-Occurrence metrics) and qualitative terms.  We compare the proposed approach with five baselines `Outpainting-SRN' \cite{wang2019wide}, `Boundless' \cite{teterwak2019boundless}, `SPGNet' \cite{song2018spg}, `SPGNet++' and partial convolutions (`PConv') \cite{liu2018image}. `PConv' \cite{liu2018image} was originally proposed for image inpainting, like in \cite{teterwak2019boundless}, we adapt it for the task of image out-painting in our setting. We also use SPGNet \cite{song2018spg} as a baseline since it also operates in semantic label space but for image inpainting task; but we adapt it for out-painting task. We create a modified version of SPGNet, SPGNet++ using \cite{park2019semantic} base generator and multiscale discriminator used in our method while retaining the exact training procedure and loss functions used in \cite{song2018spg} and use it to compare with our method. We train these baselines on our kind of input-crop (25\% of the original image).

\noindent \textbf{Evaluation Metrics} \newline
To compare the perpetual quality of the generated RGB image we use Frêchet Inception distance (FID) \cite{heusel2017gans} metric. However, since we additionally focus on generation of new objects in the extrapolated region, we also evaluate the results in semantic label space using similarity in object co-occurrence (SOCC) statistics \cite{li2019grains}.

\textbf{FID}: It is a standard metric used to calculate the fidelity of GAN generated images and provides a measure of the distance between the generated images and the real images.

\textbf{SOCC}:
The co-occurrence measure for two classes $c_a$ and $c_b$ can be calculated as the ratio of the number of times they occur together to the total number of times one of them occurs in the entire dataset. Let $N_{c_a}$ represent the frequency of a class $c_a$ in the input image, and $N_{c_{ab}}$ be the number of times there is atleast one instance of class $c_b$ present in the extrapolated region, given $c_a$ is present in the input. The probability of co-occurrence $p(c_a, c_b)$ of the two classes can be calculated as $\frac{N_{c_{ab}}}{N_{c_a}}$. The similarity in co-occurrence probability of a pair of classes between generated outputs and the training set, therefore, reflects the extent of faithful emulation of scene distribution. The similarity in co-occurrence for class $c_2$ in the output to the training set, given $c_1$ is present in the output, is defined as $s(c_a, c_b) = 1 - |p_{train}(c_a, c_b) - p_{gen}(c_a, c_b)|$. The closer this score is to 1, the greater is the similarity between the outputs of the model and the training set images.

\begin{table}[h]
\begin{center}
\begin{tabular}{|l|c|c|}
\hline
Method & Cityscapes & ADE20K\\
\hline\hline
Pconv & 86.82 & 147.14\\
Boundless & 77.36 & 136.98\\
Outpainting-SRN & 66.89 & 140.98\\
SPGNet & 83.84 & 197.69\\
SPGNet++ & 52.14 & 97.23\\
\textbf{Ours} & \textbf{47.67} & \textbf{90.45}\\
\hline
\end{tabular}
\end{center}
\vspace{-5pt}
\caption{\textbf{Results}: FID scores (lower is better) for our method vs the baselines on Cityscapes and ADE20K-bedroom dataset.}
\label{table:fid}
\vspace{-10pt}
\end{table}

\subsection{Qualitative performance}

In figure \ref{fig:stagewise_result}, we show the various stage-wise results of our pipeline. In figures \ref{fig:baselines_city} and figure \ref{fig:baselines_ade}, we compare our results with the baselines for the Cityscapes and ADE20K dataset respectively. It can be seen that our method not only extrapolates the existing objects, ensuring texture and structural consistency but is also capable of adding very precise novel objects, which the baselines fail to do. Almost all the baseline methods that operate on RGB space (except SPGNet) have trivial block like patches in the extrapolated region which is more noticeable in ADE20K dataset results.

\vspace{-5pt}
\subsection{Quantitative performance}
Table \ref{table:fid} shows the FID scores of our method compared to the baselines on the Cityscapes and ADE20K-bedroom and validation datasets. We outperform all the baselines by very significant margins.

Tables \ref{table:socc_ade20k} and \ref{table:socc_cityscapes},  show the SOCC scores for different pairs of object classes in both the datasets. Our method is able to generate results that consistently resemble the object co-occurrence statistics for most class pairs in the datasets.

\vspace{-5pt}
\section{Ablation study}
In this section, we discuss the importance of individual components of the proposed approaches in our pipeline. Table \ref{table:ablation} shows the FID scores for variants of our method on cityscapes dataset. 

\begin{table}[]
\begin{center}
    \begin{tabular}{|c|c|}
    \hline Method & FID \\ \hline
    SPGNet++ & 52.14\\
    Our (base w/o $D_{patch}$ w/o IaCN )  &  48.77\\
    Our (w/ $D_{patch}$ w/o IaCN) & 48.72\\ 
    Our (w/o $D_{patch}$ w/ IaCN) & 47.76\\
    Our (final) & \textbf{47.67}\\ \hline
\end{tabular}
\end{center}
\vspace{-5pt}
\caption{\textbf{Ablation}: FID ablation study on cityscapes dataset}
\label{table:ablation}
\vspace{-10pt}
\end{table}

\subsection{Use of Boundary maps as an extra channel for semantic extrapolation}
\begin{figure*}
\begin{center}
\includegraphics[width=0.9\linewidth]{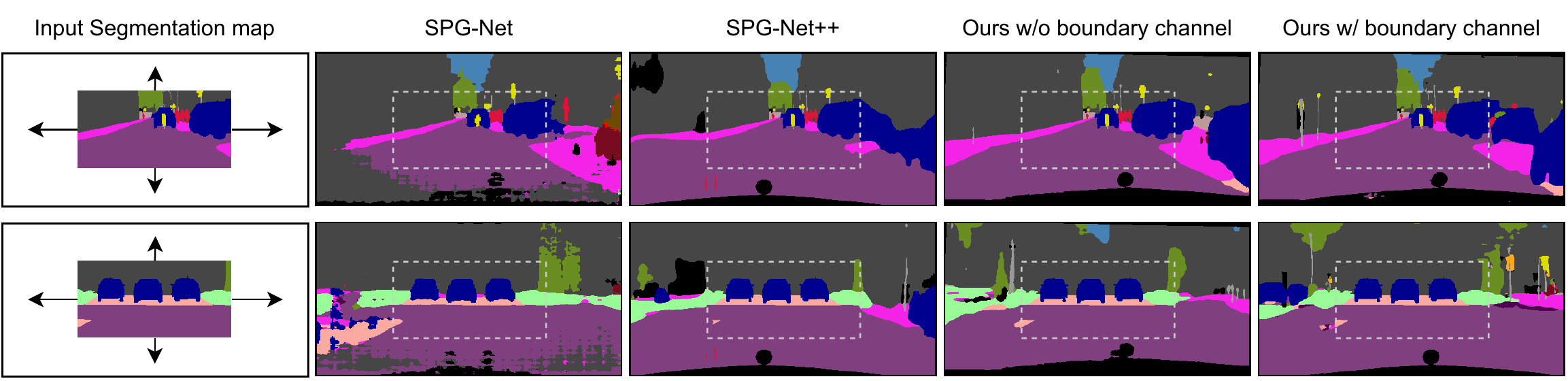}
\end{center}
    \vspace{-5pt}
  \caption{\textbf{Semantic label extrapolation ablation}: Shapes of the extrapolated or newly synthesized instances are more realistic when boundary channel is used.}
\label{fig:semantic_compare}
\vspace{-8pt}
\end{figure*}

As discussed in Section \ref{semantic}, we use the semantic class boundary map to enforce the object shape information into the network during the training time. In Figure \ref{fig:semantic_compare}, while other approaches (SPG-Net and SPG-Net++) resulted in blobs representing newly generated instances, the shapes of those instances are much better when enforced with boundary maps during training.

\subsection{Use of Panoptic label maps (Stage-3)}
\begin{figure}
\begin{center}
\includegraphics[width=0.8\linewidth]{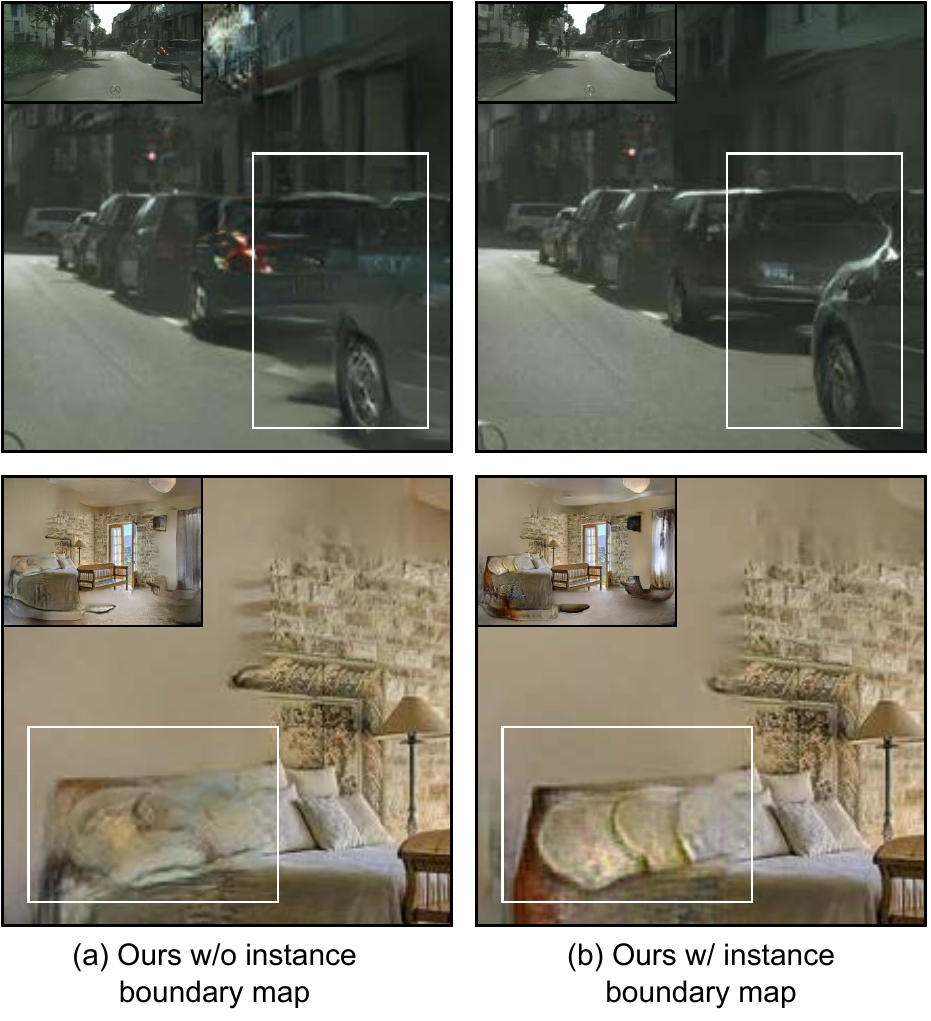}
\end{center}
    \vspace{-5pt}
  \caption{\textbf{Instance boundary map ablation}: Crisp boundaries between instances are clearly visible in the region highlighted with white box between (a) 2 cars (Cityscapes) (b) pillows on the bed (ADE20K), when boundary maps derived from the estimated panoptic label maps are used. We show the complete extrapolated image at the top-left of each image.}
\label{fig:instance_ablation}
\vspace{-7pt}
\end{figure}

We use the synthesized panoptic label maps to obtain (i) boundary maps and (ii) the feature maps for IaCN module. The boundary maps are used to generate crisp boundaries between instances of the same class (Figure \ref{fig:instance_ablation}).

\begin{figure}[t]
\begin{center}
\vspace{-4pt}
\includegraphics[width=0.8\linewidth]{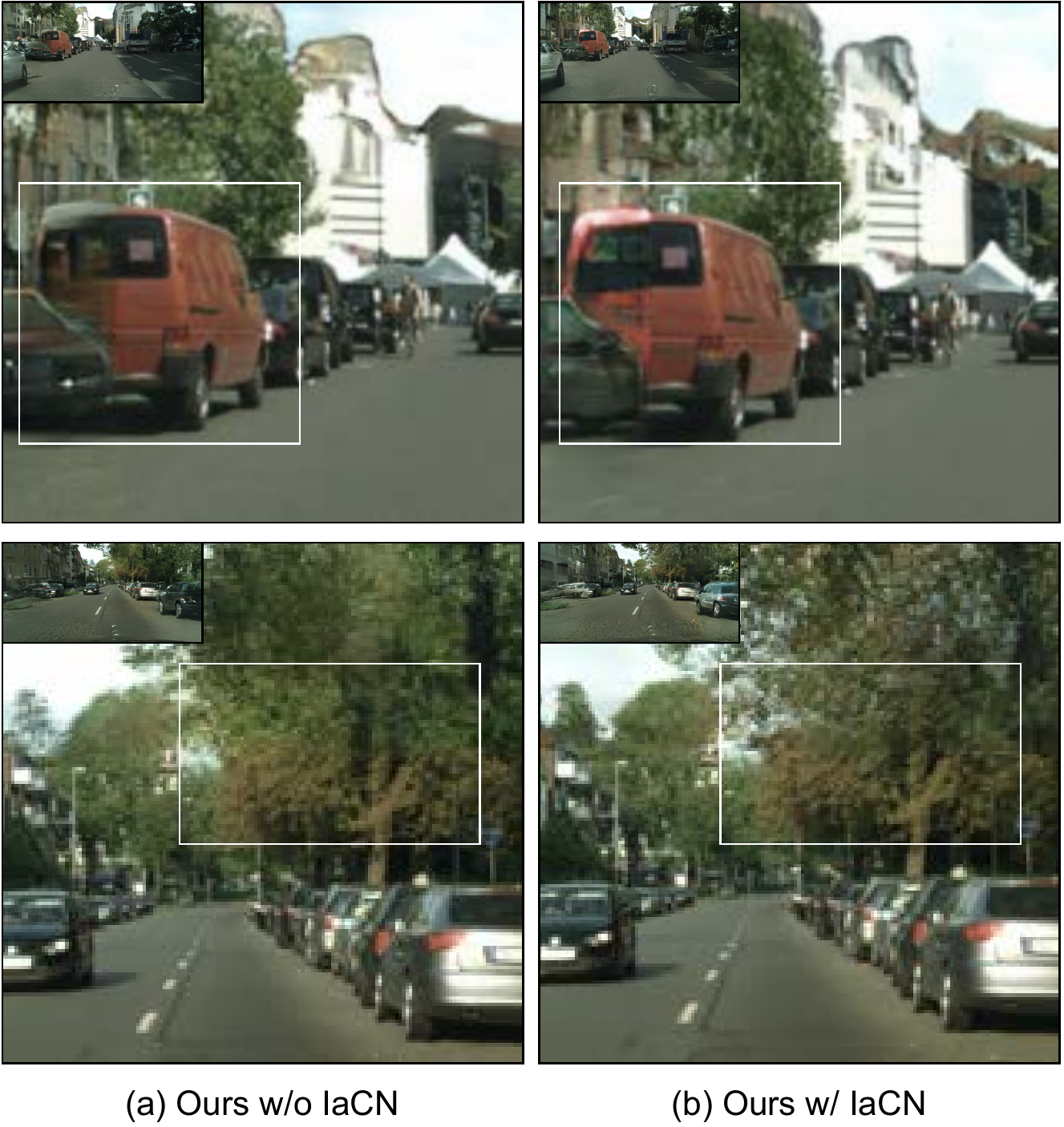}
\end{center}
    \vspace{-5pt}
  \caption{\textbf{IaCN ablation}: Texture is consistently transferred for extrapolated instances in the region highlighted with white box in (a) red van (`things' class) (Cityscapes) (b) yellowish-green tree (`stuff' class) (Cityscapes), when IaCN is used. The complete extrapolated image is present at top-left of each image.}
\label{fig:iacn_ablation}
\vspace{-5pt}
\end{figure}

\noindent \textbf{Use of IaCN module} \newline
As discussed in Section \ref{sec:iacn}, we use the feature maps generated by IaCN to preserve the texture of the extrapolated instances. Figure \ref{fig:iacn_ablation} shows that the relative texture of the extrapolated instances for both `things' (van) and `stuff' (tree) class are maintained when IaCN is used.

\subsection{Use of patch co-occurrence discriminator}
\begin{figure}
\begin{center}
\includegraphics[width=0.8\linewidth]{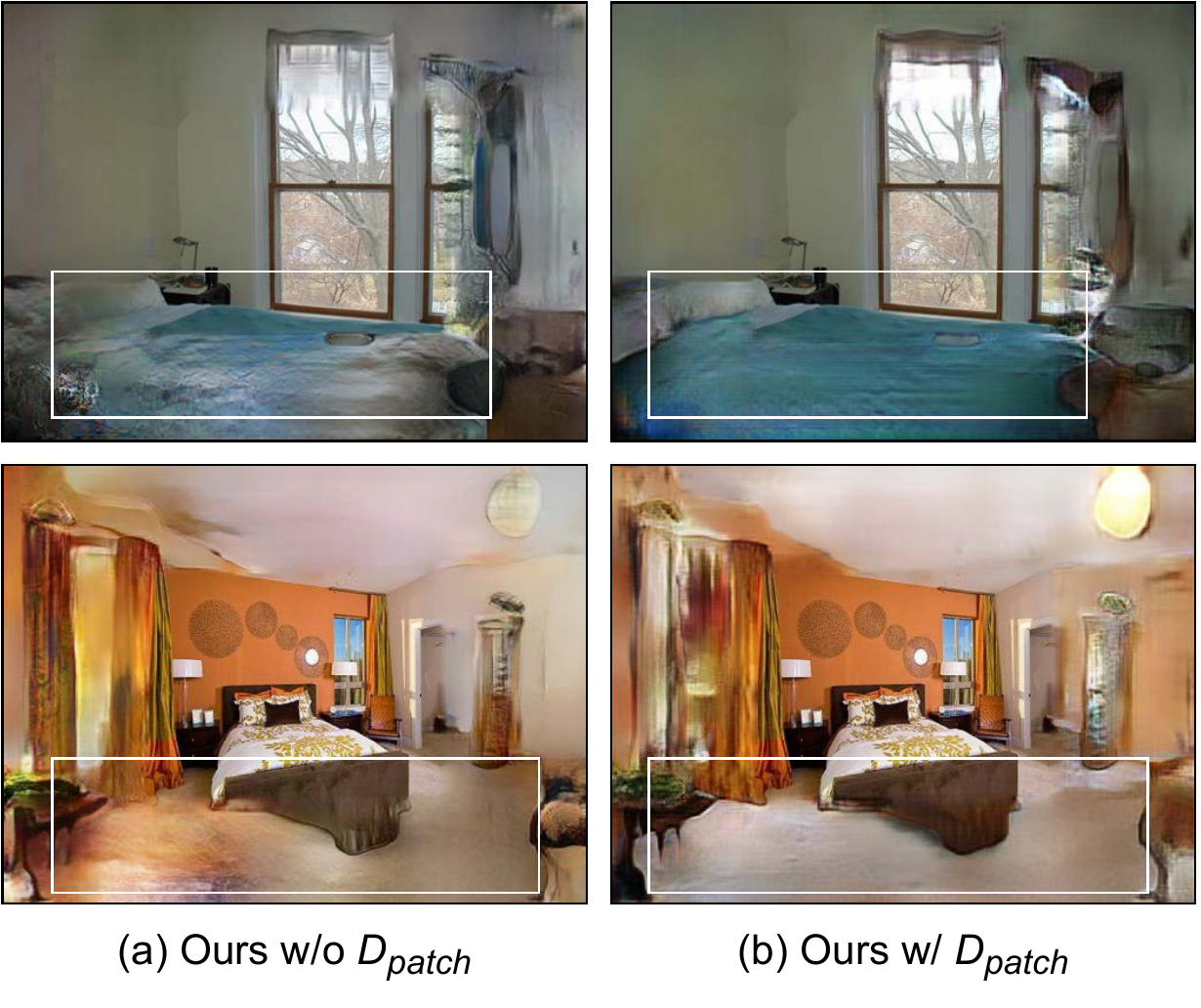}
\end{center}
    \vspace{-5pt}
  \caption{\textbf{$D_{patch}$ ablation}: Sharper images with consistent texture are produced in the region highlighted with white box in (a) blue bed-sheet (ADE20K) (b) off-white floor (ADE20K) when patch co-occurrence discriminator is used.}
\label{fig:iacn_ablation_ade}
\vspace{-15pt}
\end{figure}

Figure \ref{fig:iacn_ablation_ade} shows that sharper images are produced when patch co-occurrence discriminator is used. It also ensures much better consistent textures at the boundary of input and extrapolated regions.

\section{Discussions and Conclusion}
We propose a new solution for image extrapolation that is amenable for adding novel objects as well extending the existing objects and textures. Our solution distinguishes itself from all previous works in the image extrapolation by extrapolating the image in semantic label space. We show in the paper that this helps us achieve our objective of adding new objects. We also propose the generation of panoptic label maps from just segmentation maps, which enables us to create multiple instances of the same classes and as well allow us to have control over the instances thus created. We show in Section \ref{sec:inf_zoom} how our method can be recursively applied to generate image extrapolations of arbitrary dimensions. We hope our work encourages researchers to develop solutions for image editing in semantic label space.

\clearpage
\section*{Acknowledgement}
The authors thank Rajhans Singh for providing the processed ADE20K-bedroom subset and Richard Zhang for suggestions on writing the paper.

{\small
\bibliographystyle{ieee_fullname}
\bibliography{semie}

\begin{thebibliography}{10}\itemsep=-1pt

\bibitem{arjovsky2017wasserstein}
Martin Arjovsky, Soumith Chintala, and L{\'e}on Bottou.
\newblock Wasserstein generative adversarial networks.
\newblock In {\em International conference on machine learning}, pages
  214--223. PMLR, 2017.

\bibitem{azadi2019semantic}
Samaneh Azadi, Michael Tschannen, Eric Tzeng, Sylvain Gelly, Trevor Darrell,
  and Mario Lucic.
\newblock Semantic bottleneck scene generation.
\newblock {\em arXiv preprint arXiv:1911.11357}, 2019.

\bibitem{barnes2009patchmatch}
Connelly Barnes, Eli Shechtman, Adam Finkelstein, and Dan~B Goldman.
\newblock Patchmatch: A randomized correspondence algorithm for structural
  image editing.
\newblock {\em ACM Trans. Graph.}, 28(3):24, 2009.

\bibitem{cheng2020panoptic}
Bowen Cheng, Maxwell~D Collins, Yukun Zhu, Ting Liu, Thomas~S Huang, Hartwig
  Adam, and Liang-Chieh Chen.
\newblock Panoptic-deeplab: A simple, strong, and fast baseline for bottom-up
  panoptic segmentation.
\newblock In {\em Proceedings of the IEEE/CVF Conference on Computer Vision and
  Pattern Recognition}, pages 12475--12485, 2020.

\bibitem{cordts2016cityscapes}
Marius Cordts, Mohamed Omran, Sebastian Ramos, Timo Rehfeld, Markus Enzweiler,
  Rodrigo Benenson, Uwe Franke, Stefan Roth, and Bernt Schiele.
\newblock The cityscapes dataset for semantic urban scene understanding.
\newblock In {\em Proceedings of the IEEE conference on computer vision and
  pattern recognition}, pages 3213--3223, 2016.

\bibitem{DB16c}
A. Dosovitskiy and T. Brox.
\newblock Generating images with perceptual similarity metrics based on deep
  networks.
\newblock In {\em Advances in Neural Information Processing Systems (NIPS)},
  2016.

\bibitem{durugkar2016generative}
Ishan Durugkar, Ian Gemp, and Sridhar Mahadevan.
\newblock Generative multi-adversarial networks.
\newblock {\em arXiv preprint arXiv:1611.01673}, 2016.

\bibitem{efros2001image}
Alexei~A Efros and William~T Freeman.
\newblock Image quilting for texture synthesis and transfer.
\newblock In {\em Proceedings of the 28th annual conference on Computer
  graphics and interactive techniques}, pages 341--346, 2001.

\bibitem{790383}
A.~A. {Efros} and T.~K. {Leung}.
\newblock Texture synthesis by non-parametric sampling.
\newblock In {\em Proceedings of the Seventh IEEE International Conference on
  Computer Vision}, volume~2, pages 1033--1038 vol.2, 1999.

\bibitem{goodfellow2014generative}
Ian Goodfellow, Jean Pouget-Abadie, Mehdi Mirza, Bing Xu, David Warde-Farley,
  Sherjil Ozair, Aaron Courville, and Yoshua Bengio.
\newblock Generative adversarial nets.
\newblock In {\em Advances in neural information processing systems}, pages
  2672--2680, 2014.

\bibitem{guo2020spiral}
Dongsheng Guo, Hongzhi Liu, Haoru Zhao, Yunhao Cheng, Qingwei Song, Zhaorui Gu,
  Haiyong Zheng, and Bing Zheng.
\newblock Spiral generative network for image extrapolation.
\newblock In {\em European Conference on Computer Vision}, pages 701--717.
  Springer, 2020.

\bibitem{guo2020spiralnet}
Dongsheng Guo, Hongzhi Liu, Haoru Zhao, Yunhao Cheng, Qingwei Song, Zhaorui Gu,
  Haiyong Zheng, and Bing Zheng.
\newblock Spiral generative network for image extrapolation.
\newblock In {\em The European Conference on Computer Vision (ECCV)}, pages
  701--717, 2020.

\bibitem{heusel2017gans}
Martin Heusel, Hubert Ramsauer, Thomas Unterthiner, Bernhard Nessler, and Sepp
  Hochreiter.
\newblock Gans trained by a two time-scale update rule converge to a local nash
  equilibrium.
\newblock In {\em Advances in neural information processing systems}, pages
  6626--6637, 2017.

\bibitem{hong2018learning}
Seunghoon Hong, Xinchen Yan, Thomas Huang, and Honglak Lee.
\newblock Learning hierarchical semantic image manipulation through structured
  representations.
\newblock {\em arXiv preprint arXiv:1808.07535}, 2018.

\bibitem{isola2017image}
Phillip Isola, Jun-Yan Zhu, Tinghui Zhou, and Alexei~A Efros.
\newblock Image-to-image translation with conditional adversarial networks.
\newblock In {\em Proceedings of the IEEE conference on computer vision and
  pattern recognition}, pages 1125--1134, 2017.

\bibitem{johnson2016perceptual}
Justin Johnson, Alexandre Alahi, and Li Fei-Fei.
\newblock Perceptual losses for real-time style transfer and super-resolution.
\newblock In {\em European conference on computer vision}, pages 694--711.
  Springer, 2016.

\bibitem{karras2020analyzing}
Tero Karras, Samuli Laine, Miika Aittala, Janne Hellsten, Jaakko Lehtinen, and
  Timo Aila.
\newblock Analyzing and improving the image quality of stylegan.
\newblock In {\em Proceedings of the IEEE/CVF Conference on Computer Vision and
  Pattern Recognition}, pages 8110--8119, 2020.

\bibitem{kingma2014adam}
Diederik~P Kingma and Jimmy Ba.
\newblock Adam: A method for stochastic optimization.
\newblock {\em arXiv preprint arXiv:1412.6980}, 2014.

\bibitem{kingma2013auto}
Diederik~P Kingma and Max Welling.
\newblock Auto-encoding variational bayes.
\newblock {\em arXiv preprint arXiv:1312.6114}, 2013.

\bibitem{kirillov2019panoptic}
Alexander Kirillov, Kaiming He, Ross Girshick, Carsten Rother, and Piotr
  Doll{\'a}r.
\newblock Panoptic segmentation.
\newblock In {\em Proceedings of the IEEE/CVF Conference on Computer Vision and
  Pattern Recognition}, pages 9404--9413, 2019.

\bibitem{kulkarni2020hallucinet}
Kuldeep Kulkarni, Tejas Gokhale, Rajhans Singh, Pavan Turaga, and Aswin
  Sankaranarayanan.
\newblock Halluci-net: Scene completion by exploiting object co-occurrence
  relationships.
\newblock {\em arXiv preprint arXiv:2004.08614}, 2020.

\bibitem{lee2018context}
Donghoon Lee, Sifei Liu, Jinwei Gu, Ming-Yu Liu, Ming-Hsuan Yang, and Jan
  Kautz.
\newblock Context-aware synthesis and placement of object instances.
\newblock {\em arXiv preprint arXiv:1812.02350}, 2018.

\bibitem{li2019grains}
Manyi Li, Akshay~Gadi Patil, Kai Xu, Siddhartha Chaudhuri, Owais Khan, Ariel
  Shamir, Changhe Tu, Baoquan Chen, Daniel Cohen-Or, and Hao Zhang.
\newblock Grains: Generative recursive autoencoders for indoor scenes.
\newblock {\em ACM Transactions on Graphics (TOG)}, 38(2):1--16, 2019.

\bibitem{lim2017geometric}
Jae~Hyun Lim and Jong~Chul Ye.
\newblock Geometric gan.
\newblock {\em arXiv preprint arXiv:1705.02894}, 2017.

\bibitem{lin2017focal}
Tsung-Yi Lin, Priya Goyal, Ross Girshick, Kaiming He, and Piotr Doll{\'a}r.
\newblock Focal loss for dense object detection.
\newblock In {\em Proceedings of the IEEE international conference on computer
  vision}, pages 2980--2988, 2017.

\bibitem{liu2018image}
Guilin Liu, Fitsum~A Reda, Kevin~J Shih, Ting-Chun Wang, Andrew Tao, and Bryan
  Catanzaro.
\newblock Image inpainting for irregular holes using partial convolutions.
\newblock In {\em Proceedings of the European Conference on Computer Vision
  (ECCV)}, pages 85--100, 2018.

\bibitem{liu2019learning}
Xihui Liu, Guojun Yin, Jing Shao, Xiaogang Wang, et~al.
\newblock Learning to predict layout-to-image conditional convolutions for
  semantic image synthesis.
\newblock In {\em Advances in Neural Information Processing Systems}, pages
  570--580, 2019.

\bibitem{mao2017least}
Xudong Mao, Qing Li, Haoran Xie, Raymond~YK Lau, Zhen Wang, and Stephen
  Paul~Smolley.
\newblock Least squares generative adversarial networks.
\newblock In {\em Proceedings of the IEEE international conference on computer
  vision}, pages 2794--2802, 2017.

\bibitem{mescheder2018training}
Lars Mescheder, Andreas Geiger, and Sebastian Nowozin.
\newblock Which training methods for gans do actually converge?
\newblock In {\em International conference on machine learning}, pages
  3481--3490. PMLR, 2018.

\bibitem{miyato2018spectral}
Takeru Miyato, Toshiki Kataoka, Masanori Koyama, and Yuichi Yoshida.
\newblock Spectral normalization for generative adversarial networks.
\newblock {\em arXiv preprint arXiv:1802.05957}, 2018.

\bibitem{ntavelis2020sesame}
Evangelos Ntavelis, Andr{\'e}s Romero, Iason Kastanis, Luc Van~Gool, and Radu
  Timofte.
\newblock Sesame: Semantic editing of scenes by adding, manipulating or erasing
  objects.
\newblock In {\em European Conference on Computer Vision}, pages 394--411.
  Springer, 2020.

\bibitem{park2019semantic}
Taesung Park, Ming-Yu Liu, Ting-Chun Wang, and Jun-Yan Zhu.
\newblock Semantic image synthesis with spatially-adaptive normalization.
\newblock In {\em Proceedings of the IEEE Conference on Computer Vision and
  Pattern Recognition}, pages 2337--2346, 2019.

\bibitem{park2020swapping}
Taesung Park, Jun-Yan Zhu, Oliver Wang, Jingwan Lu, Eli Shechtman, Alexei
  Efros, and Richard Zhang.
\newblock Swapping autoencoder for deep image manipulation.
\newblock {\em Advances in Neural Information Processing Systems}, 33, 2020.

\bibitem{simonyan2015deep}
Karen Simonyan and Andrew Zisserman.
\newblock Very deep convolutional networks for large-scale image recognition.
\newblock {\em arXiv preprint arXiv:1409.1556}, 2014.

\bibitem{song2018spg}
Yuhang Song, Chao Yang, Yeji Shen, Peng Wang, Qin Huang, and C-C~Jay Kuo.
\newblock Spg-net: Segmentation prediction and guidance network for image
  inpainting.
\newblock {\em arXiv preprint arXiv:1805.03356}, 2018.

\bibitem{tao2020hierarchical}
Andrew Tao, Karan Sapra, and Bryan Catanzaro.
\newblock Hierarchical multi-scale attention for semantic segmentation.
\newblock {\em arXiv preprint arXiv:2005.10821}, 2020.

\bibitem{teterwak2019boundless}
Piotr Teterwak, Aaron Sarna, Dilip Krishnan, Aaron Maschinot, David Belanger,
  Ce Liu, and William~T Freeman.
\newblock Boundless: Generative adversarial networks for image extension.
\newblock In {\em Proceedings of the IEEE International Conference on Computer
  Vision}, pages 10521--10530, 2019.

\bibitem{wang2018high}
Ting-Chun Wang, Ming-Yu Liu, Jun-Yan Zhu, Andrew Tao, Jan Kautz, and Bryan
  Catanzaro.
\newblock High-resolution image synthesis and semantic manipulation with
  conditional gans.
\newblock In {\em Proceedings of the IEEE conference on computer vision and
  pattern recognition}, pages 8798--8807, 2018.

\bibitem{wang2019wide}
Yi Wang, Xin Tao, Xiaoyong Shen, and Jiaya Jia.
\newblock Wide-context semantic image extrapolation.
\newblock In {\em Proceedings of the IEEE Conference on Computer Vision and
  Pattern Recognition}, pages 1399--1408, 2019.

\bibitem{wang2020sketchguided}
Yaxiong Wang, Yunchao Wei, Xueming Qian, Li Zhu, and Yi Yang.
\newblock Sketch-guided scenery image outpainting.
\newblock {\em arXiv preprint arXiv:2006.09788}, 2020.

\bibitem{xian2018texturegan}
Wenqi Xian, Patsorn Sangkloy, Varun Agrawal, Amit Raj, Jingwan Lu, Chen Fang,
  Fisher Yu, and James Hays.
\newblock Texturegan: Controlling deep image synthesis with texture patches.
\newblock In {\em Proceedings of the IEEE Conference on Computer Vision and
  Pattern Recognition}, pages 8456--8465, 2018.

\bibitem{yang2019very}
Zongxin Yang, Jian Dong, Ping Liu, Yi Yang, and Shuicheng Yan.
\newblock Very long natural scenery image prediction by outpainting.
\newblock In {\em Proceedings of the IEEE International Conference on Computer
  Vision}, pages 10561--10570, 2019.

\bibitem{Yeh_2017_CVPR}
Raymond~A. Yeh, Chen Chen, Teck Yian~Lim, Alexander~G. Schwing, Mark
  Hasegawa-Johnson, and Minh~N. Do.
\newblock Semantic image inpainting with deep generative models.
\newblock In {\em Proceedings of the IEEE Conference on Computer Vision and
  Pattern Recognition (CVPR)}, July 2017.

\bibitem{yu2020context}
Changqian Yu, Jingbo Wang, Changxin Gao, Gang Yu, Chunhua Shen, and Nong Sang.
\newblock Context prior for scene segmentation.
\newblock In {\em Proceedings of the IEEE/CVF Conference on Computer Vision and
  Pattern Recognition}, pages 12416--12425, 2020.

\bibitem{yu2018generative}
Jiahui Yu, Zhe Lin, Jimei Yang, Xiaohui Shen, Xin Lu, and Thomas~S Huang.
\newblock Generative image inpainting with contextual attention.
\newblock In {\em Proceedings of the IEEE conference on computer vision and
  pattern recognition}, pages 5505--5514, 2018.

\bibitem{zhang2019selfattention}
Han Zhang, Ian Goodfellow, Dimitris Metaxas, and Augustus Odena.
\newblock Self-attention generative adversarial networks.
\newblock In {\em International Conference on Machine Learning}, pages
  7354--7363. PMLR, 2019.

\bibitem{zhang2020resnest}
Hang Zhang, Chongruo Wu, Zhongyue Zhang, Yi Zhu, Zhi Zhang, Haibin Lin, Yue
  Sun, Tong He, Jonas Mueller, R Manmatha, et~al.
\newblock Resnest: Split-attention networks.
\newblock {\em arXiv preprint arXiv:2004.08955}, 2020.

\bibitem{Zhang_2020}
Xiaofeng Zhang, Feng Chen, Cailing Wang, Ming Tao, and Guo-Ping Jiang.
\newblock Sienet: Siamese expansion network for image extrapolation.
\newblock {\em IEEE Signal Processing Letters}, 27:1590–1594, 2020.

\bibitem{zhao2017pyramid}
Hengshuang Zhao, Jianping Shi, Xiaojuan Qi, Xiaogang Wang, and Jiaya Jia.
\newblock Pyramid scene parsing network.
\newblock In {\em Proceedings of the IEEE conference on computer vision and
  pattern recognition}, pages 2881--2890, 2017.

\bibitem{Zhou_2017_CVPR}
Bolei Zhou, Hang Zhao, Xavier Puig, Sanja Fidler, Adela Barriuso, and Antonio
  Torralba.
\newblock Scene parsing through ade20k dataset.
\newblock In {\em Proceedings of the IEEE Conference on Computer Vision and
  Pattern Recognition (CVPR)}, July 2017.

\end{thebibliography}
}

\clearpage
\appendix
\section{Network Architectures}
\label{sec:net_arch}
The stage 1 uses PSPnet \cite{zhao2017pyramid} to obtain the semantic label maps of the input images. The generators used in stage 2 (semantic extrapolation), stage 3 (panoptic label map generation) and stage 4 (RGB image generation) are inspired from SPADE \cite{park2019semantic} generator and consist of 9 spade residual blocks. The multiscale discriminators used in stage 2 and stage 4 are similar to that in pix2pixHD \cite{wang2018high}. The encoder of stage 4 (that forms VAE \cite{kingma2013auto} with the generator) is inspired from the encoder used in SPADE \cite{park2019semantic}. The co-occurrence patch discriminator of stage 4 is inspired from the one used in Swapping AutoEncoders \cite{park2020swapping}. Our stage 3 which converts the semantic label map into panoptic label maps \footnote{We use the term `panoptic label map' and `instance map` interchangeably} is a pure-generator network and is inspired by \cite{cheng2020panoptic}. 

\begin{figure*}
\begin{center}
\includegraphics[width=0.9\linewidth]{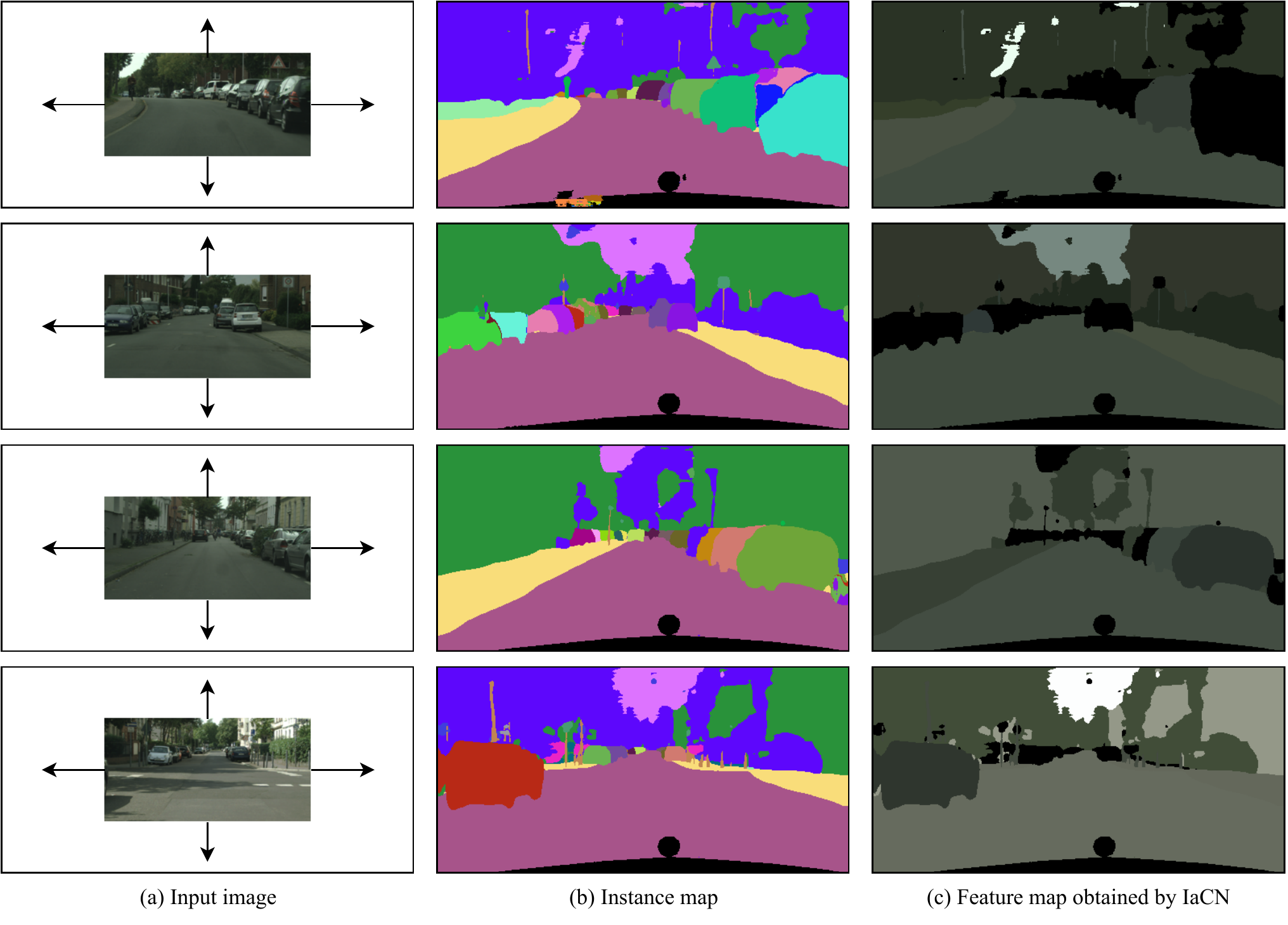}
\end{center}
  \caption{\textbf{Input-Output for IaCN}. It computes the mean colors for all the partial instances from the cropped image and just `pastes' them in the extrapolated regions of the corresponding instances. For all the other instances, it has zero value (black).}
\label{fig:iacn}
\end{figure*}

\section{Panoptic Map generation (stage 3)}
\label{sec:stage3_supp}
The stage 2 of our pipeline extrapolates the semantic label maps. However, in order to differentiate between multiple instances of the same class, we need to generate the panoptic label map from the thus extrapolated semantic label maps. In a typical panoptic label segmentation set-up, we have access to the full image. However, in our case we do not have access to the full image, but we have access to only the extrapolated semantic label map which is obtained as output of stage 2. To obtain the panoptic label maps from outputs of stage 2, we take inspiration from \cite{cheng2020panoptic}. Panoptic-DeepLab proposes a method to convert the image into two parallel branches 1. semantic label maps and 2. pixel-wise instance center maps and $x$ and $y$ off-set maps from the instance centers. The center maps and the off-set maps thus predicted are used in conjunction with the semantic label maps to obtain the final panoptic label map.  For the panoptic segmentation branch, \cite{cheng2020panoptic} obtain class-agnostic instance centers and off-sets from the instance centers for every location. Class-agnostic centers refer to center locations for the different instances that belong to the `things' categories. In addition, for every pixel that belongs to the `things' categories, we define the x-offsets and y-offsets as $\delta x$ and $\delta y$, respectively, of that pixel location from the center of the instance the pixel belongs to. Here, instead of having the above mentioned two parallel branches, we train a network to obtain the center maps and the $x$ and $y$ offset maps from the semantically extrapolated label maps that are the outputs of stage 2. The ground-truth center maps are represented by Gaussian blobs of standard deviation of 8 pixels, centered at the instance centers. We use a simple $L_2$ loss to compute the instance center loss and $L_1$ loss to compute the offset losses. The final loss for stage 3 is the weighted sum of the center loss and the offset losses. 

During the test time, we adapt the procedure mentioned by \cite{cheng2020panoptic} to group the pixels based on the predicted centers and off-sets to form instance masks. The instance masks and the semantic label map (the input to stage 3) are combined by majority voting to obtain the panoptic label map. The thus obtained panoptic label map is used in stage 4 for our instance-aware context normalization as well as to obtain the instance boundary maps.

\section{Instance-aware Context Normalization (IaCN)}
This module takes in input the cropped RGB image and the generated panoptic label map from stage 3. The panoptic label map is used to get the partial instances. Partial instances refer to the instances which are part of the input image and need to be completed in the final out-painted image. The per-channel (for RGB) mean color for each partial instance is calculated as mean pixel value of all the pixels belonging to that instance. The feature map is obtained by copying these mean colors (per-channel) to their respective extrapolated part of the partial instances in the outside region. For all the instances that do not belong to partial instances, the feature map values are zero. Figure \ref{fig:iacn} shows some of the input-output pairs for IaCN module.

\section{Detailed Training and Testing algorithms}
\label{sec:algo}
The detailed training algorithms for stage 2 and stage 4 are given in Algorithm \ref{algo:training_algo_stage2} and \ref{algo:training_algo_stage4}.

To train stage 2 (as shown in Algorithm \ref{algo:training_algo_stage2}), we use the ground truth segmentation map $S_{gt}$, ground truth panoptic map $P_{gt}$ and cropped segmentation map $S_{gt}^{c}$ (obtained from stage 1) as the inputs. We obtain segmentation map, $S_{gt}^{z}$, of desired resolution by zero padding $S_{gt}^{c}$. An instance boundary map, $B$ is created from $P_{gt}$. Extrapolated segmentation map $S_{pog}$ is generated using the generator $G^2$ which has an extra output channel (apart from the input classes) for the boundary map. The multiscale discriminator, $D_\text{multiscale}^2$ distinguishes between the generated segmentation map ($S_{pog}$) and the ground truth segmentation map ($S_{gt}$). The model tries to minimize the train objective function for semantic label map extrapolation (Section \ref{sec:stage2_obj}). Finally, the parameters of $G^{2}$, $D_\text{multiscale}^{2}$ are updated accordingly.

\begin{algorithm}
\DontPrintSemicolon
\KwInput{ \\ Ground Truth Segmentation Map: $S_{gt} \in \{0,1\}^{2h \times 2w \times c}$,\\
         Ground Truth Panoptic Map: $P_{gt} \in \mathbb{R}^{2h \times 2w \times 1} $,\\
         Cropped Segmentation Map: $S_{gt}^{c} \in \{0,1\}^{h \times w \times c}$}

Generate $S_{gt}^{z} \in \{0,1\}^{2h \times 2w \times c}$ by zero-padding $S_{gt}^{c}$\;
Generate Boundary Map $B \longleftarrow GetBoundary(P_{gt})$\;
\For{epoch in maxEpochs} {
    $S_{pog} \longleftarrow G^{2}(L_1)$\;
    $D_{\text{multiscale}}^{2}$ distinguishes between $S_{pog}$ and $S_{gt} $\;
    Minimize the objective function\;
    Update the parameters of $G^{2}, D_{\text{multiscale}}^{2}$
}
\Return $G^{2}$

\caption{Training algorithm for stage 2}
\label{algo:training_algo_stage2}
\end{algorithm}

To train stage 4 (as shown in Algorithm \ref{algo:training_algo_stage4}), we obtain $X_{com}$ by concatenating ground truth segmentation map ($S_{gt}$), input image ($X_{gt}^{c}$), boundary map obtained from ground truth instance map and the feature map obtained using IaCN module. The out-painted image ($Y$) is generated using the generator of stage 4 $G^{4}$, which takes in $X_{com}$ and the encoded input image ($E^{4}(X_{gt}^{c})$). The multiscale discriminator $D_\text{multiscale}^{4}$ tries to distinguish between the generated image ($Y$) and the ground truth image ($X_{gt}$). The co-occurrence patch discriminator ($D_\text{patch}^{4}$) tries to distinguish between fake patch ($crop(Y)$) and real patch ($crop(X_{gt})$) obtained from the fake image and real images respectively. For each pair of fake patch and real patch, the patch discriminator takes 4 reference patches ($crops(X_{gt})$) from the real image. We take a total of 4 fake patch, real patch and reference patches combinations for one image. All the patches are of size $64 \times 64$. We, then, minimize the final objective function (Section \ref{sec:stage4_obj}) to update the parameters of $G^{4}, E^{4}, D_\text{multiscale}^{4}$ and $D_\text{patch}^{4}$.

\begin{algorithm}
\DontPrintSemicolon
\KwInput{\\Cropped Image: $X_{gt}^{c} \in \mathbb{R}^{h \times w \times 3} $,\\
         Ground Truth Image: $X_{gt} \in \mathbb{R}^{2h \times 2w \times 3} $,\\
         Ground Truth Segmentation Map: $S_{gt} \in \{0,1\}^{2h \times 2w \times c}$,\\
         Ground Truth Panoptic Map: $P_{gt} \in \mathbb{R}^{2h \times 2w \times 1}$}

$X_{com} \longleftarrow S_{gt} \oplus X_{gt}^{c} \oplus GetBoundary(P_{gt}) \oplus IaCN(X_{gt}^{c}, P_{gt})$
\For{epoch in maxEpochs} {
    $Y \longleftarrow G^{4}(X_{com}, E^{4}(X_{gt}^{c}))$\;
    $D_{\text{multiscale}}^{4}$ distinguishes between $Y$ and $X_{gt}$\;
    $D_{\text{patch}}^{4}$ distinguishes between $crop(Y)$ and $crop(X_{gt})$, taking $crops(X_{gt})$ as the ref. patches\;
    Minimize the objective function\;
    Update the parameters of $G^{4}, E^{4}, D_{\text{multiscale}}^{4}, D_{\text{patch}}^{4}$
}
\Return $E^{4}, G^{4}$

\caption{Training algorithm for stage 4}
\label{algo:training_algo_stage4}
\end{algorithm}

The detailed testing algorithm is given in Algorithm \ref{algo:testing_algo}. The semantic label map $S^{c}$ corresponding to input image $X$ is obtained from PSPnet \cite{zhao2017pyramid} in stage 1. The extrapolated semantic label map $S_{pog}$ is generated from generator $G^{2}$ of stage 2. $S_{pog}$ is fed into stage 3 (Panoptic Label Generatior) to get panoptic map $P'$. $S_{pog}, X_{gt}^{c}$, boundary map obtained from $P'$ and output of $IaCN$ are concatenated into $X'_{com}$, which is given as input to generator $G^{4}$ in stage 4 to generate the extrapolated RGB image $Y$.

\begin{algorithm}
\DontPrintSemicolon
\KwInput{Image: $X \in \mathbb{R}^{h \times w \times 3} $}
\KwOutput{Outpainted Image: $Y \in \mathbb{R}^{2h \times 2w \times 3} $}
$S^{c} \longleftarrow \textbf{PSPNet}(X_{gt}^{c})$ \tcp*{Stage 1}
$S_{pog} \longleftarrow G^{2}(S^{c})$ \tcp*{Stage 2}
$P' \longleftarrow \textbf{PanopticLabelMap}(S_{pog})$ \tcp*{Stage 3}
$X'_{com} \longleftarrow S_{pog} \oplus X \oplus GetBoundary(P') \oplus IaCN(X, P')$

$Y \longleftarrow G^{4}(X'_{com}, E^4(X))$ \tcp*{Stage 4}

\caption{Testing algorithm}
\label{algo:testing_algo}
\end{algorithm}

\vspace{-5pt}
\section{Objective Functions}

This section describes all the loss functions used in different stages of training in detail. All the notations for different variables is the same as described in Section \ref{sec:algo}.

\subsection{Stage 2}
\label{sec:stage2_obj}
The objective function for stage 2 includes 4 losses: GAN loss, discriminator feature matching loss, focal loss and cross-entropy loss. 

\textbf{GAN loss:} $S_{gt}^{c}$ is the semantic label map corresponding to input image, $S_{gt}$ is the corresponding extrapolated ground truth semantic label map. $B$ is the ground truth boundary map obtained from ground truth instance map $P_{gt}$. $S_{com}$ is channel-wise contatenation of $S_{gt}$ and $B$. $S_{pog}$ is the combined extrapolated sematic label map and boundary map  synthesised by $G^{2}$. We replace GAN hinge loss used in \cite{park2019semantic} with least square loss ($\mathbf{\mathcal{L}}_{GAN}$).

\textbf{Feature matching loss:} For stability in GAN training, we use feature matching loss ($\mathbf{\mathcal{L}}_{FM}$) which is defined as,
\begin{equation}
\label{eqn:stage2_fm}
\begin{split}
     \mathcal{L}_{FM} = \sum_{i}\frac{1}{N_{i}}[\|D^{2(i)}(S_{com}) - D^{2(i)}(S_{pog})\|_{1}]
\end{split}
\end{equation}
where $D^{2(i)}$ represents $i-$th layer of discriminator $D^{2}$ with $N_{i}$ elements.

\textbf{Cross-entropy loss:}
We apply a cross-entropy loss ($\mathbf{\mathcal{L}}_{CE}$) on input \textit{$Y''$}YY and $Y$ defined as,
\begin{equation}
\label{eqn:stage2_ce_1}
\begin{split}
     l(z,y) = -(y log(z) + (1-y) log(1-z)
\end{split}
\end{equation}

\begin{equation}
\label{eqn:stage2_ce}
\begin{split}
     \mathcal{L}_{CE} = \sum_{h,w,c} l(z,y)
\end{split}
\end{equation}
where $l(z,y)$ defines element wise cross-entropy loss for particular $(h,w,c)$,  and $\mathbf{\mathcal{L}}_{CE}$ is the total cross-entropy loss. y and z are the ground-truth and generated values at each $(h,w,c)$ location in $S_{com}$ and $S_{pog}$ respectively.

\textbf{Focal loss:} To better account for representation of rare semantic classes in the generated semantic label map, we use focal loss \cite{lin2017focal} ($\mathbf{\mathcal{L}}_{FL}$) defined as,
\begin{equation}
\label{eqn:stage2_fl}
\begin{split}
     \mathcal{L}_{FL} = \sum_{h,w,c} l(z,y) \times (1-z)^{\gamma}
\end{split}
\end{equation}

\textbf{Training objective:} The overall training objective for this stage, hence, is $\mathbf{\mathcal{L}}_{2}$,
\begin{equation}
\label{eqn:stage2}
\begin{split}
     \mathcal{L}_{2} = \min\limits_{G^2} ( &\mathcal{L}_{GAN} + \lambda_{FM}\mathcal{L}_{FM} + \lambda_{CE}\mathcal{L}_{CE} \\& + \lambda_{FL}\mathcal{L}_{FL})
\end{split}
\end{equation}

\subsection{Stage 4}
\label{sec:stage4_obj}
The objective function for stage 4 includes 5 losses: GAN loss, discriminator feature matching loss, perceptual loss, KL-Divergence loss and Patch co-occurrence discriminator loss. 

\textbf{GAN loss:} $X_{gt}^{c}$ is the input image. $X_{gt}$ is the extrapolated ground truth image. $S_{gt}$ is the ground truth semantic label map and $P_{gt}$ is the ground truth instance map corresponding to $S_{gt}$. $X_{com}$ is channel-wise concatenation of $X_{gt}^{c}$, $S_{gt}$, boundary map obtained from $P_{gt}$ and output of $IaCN$ module. $Y$ is the extrapolated RGB image synthesised by $G^{4}$. For GAN loss ($\mathbf{\mathcal{L}}_{GAN}$) we use hinge loss similar to \cite{park2019semantic}

\textbf{Feature matching loss:} For stability in GAN training we use feature matching loss ($\mathbf{\mathcal{L}}_{FM}$) which is defined as,
\begin{equation}
\label{eqn:stage4_fm}
\begin{split}
     \mathcal{L}_{FM} = \sum_{i}\frac{1}{N_{i}}[\|D^{4(i)}(X_{gt}^{c}) - D^{4(i)}(Y)\|_{1}]
\end{split}
\end{equation}
where $D^{4(i)}$ represents $i-$th layer of discriminator $D^{4}$ with $N_{i}$ elements.

\textbf{Perceptual loss:} We use VGG19 as feature extractor to minimize the $\mathbf{L1}$ loss between features extracted for $Z'$ and $Y$. Particularly, we define perceptual loss  ($\mathbf{\mathcal{L}}_{VGG}$) as,

\begin{equation}
\label{eqn:stage4_vgg}
\begin{split}
     \mathcal{L}_{VGG} = \sum_{i}\frac{1}{N_{i}}[\|\Phi^{(i)}(X_{gt}^{c}) - \Phi^{(i)}(Y)\|_{1}]
\end{split}
\end{equation}
where $\Phi^{i}$ denotes $i-$th layer of VGG19 network.

\textbf{KL divergence loss:} We use KLD loss ($\mathbf{\mathcal{L}}_{KLD}$), similar to \cite{park2019semantic} to train the VAE (defined in Section 3.4 in main paper). It is defined as,
\begin{equation}
\label{eqn:stage2_kld}
\begin{split}
     \mathcal{L}_{KLD} = \mathbb{D_{KL}}(q(z|x)||p(z))
\end{split}
\end{equation}
where $q$ is variation distribution and $p(z)$ is a standard Gaussian distribution.

\begin{figure}
\begin{center}
\includegraphics[width=\linewidth]{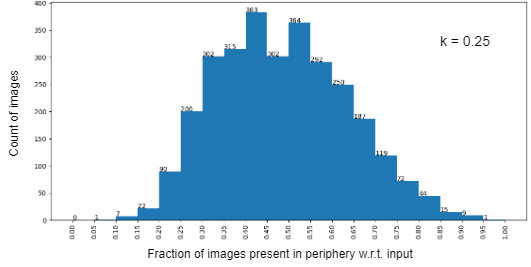}  
\includegraphics[width=\linewidth]{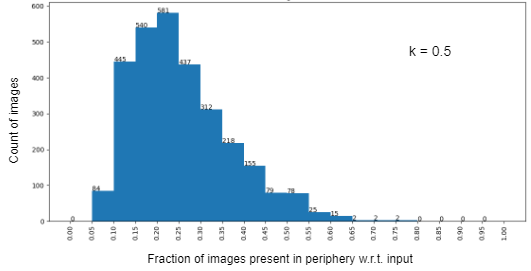}  
\includegraphics[width=\linewidth]{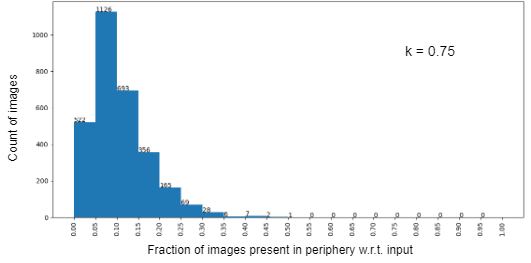}  
\end{center}
\caption{Fraction/total number of images in the cityscapes training set where the ratio of number of instances outside the input crop to the number of instances inside the input crop. For $k=0.25$, the peak occurs around 50\%, for $k=0.5$, the peak occurs around 25\% and for $k=0.75$, the peak occurs around 10\%.}
\label{fig:crop_analysis}
\end{figure}

\textbf{Patch Co-Occurrence discriminator loss:} In addition to multiscale discriminator $D^{4}$, we also use a patch co-occurrence discriminator (described in detail in Section 3.4 of main paper). The patch co-occurrence discriminator loss $\mathcal{L}_{CooccurGAN}$ is defined as,

\begin{equation}
\label{eq:swap}
\begin{split}
    &\mathcal{L}_{CooccurGAN} = \\ &\mathbb{E}_{Y,X_{gt}^{c}}[-log(D^{4}_{patch}(crop(G(Y)), crop(X_{gt}^{c}), crops(X_{gt}^{c})))]
\end{split}
\end{equation}
where $crop(X_{gt}^{c})$ function takes a random patch of $64 \times 64$ from image $X_{gt}^{c}$ and $crops(X_{gt}^{c})$ takes 4 random patches from image $X_{gt}^{c}$, which serve as the reference patches.

\textbf{Training objective:}
The overall training objective for this stage, hence, is $\mathbf{\mathcal{L}}_{4}$,
\begin{equation}
\label{eqn:stage4}
\begin{split}
     \mathcal{L}_{4} = \min\limits_{G^{4}} ( &\mathcal{L}_{GAN} +  \lambda_{FM}\mathcal{L}_{FM} + \lambda_{VGG}\mathcal{L}_{VGG} \\& + \lambda_{VGG}\mathcal{L}_{VGG}+ \mathcal{L}_{CooccurGAN})
\end{split}
\end{equation}

\section{Additional Implementation details}
We trained and tested our models on Cityscapes \cite{cordts2016cityscapes} and ADE-20K bedroom subset \cite{Zhou_2017_CVPR} datasets. For stage 2 we used a batch size of 8, while for stage 4 we used a batch size of 16. All the experiments were run on 4 16GB Nvidia Tesla V100 GPUs. Both the datasets were trained for 200 epochs. We used the TTUR \cite{heusel2017gans} update rule.

\section{Different crop analysis}
We analyzed different crop sizes on Cityscapes dataset. For given images $X_0 (\in \mathbb{R}^{h\times w\times c})$ in the dataset, we tried 3 different crop ratios $k=\{0.25, 0.5, 0.75\}$. We, thus, obtain the cropped images $X (\in \mathbb{R}^{k.h \times k.w \times c})$. We calculated the number of instances outside the crop region with respect to those inside. If the number of instances in the region to be extrapolated is very high as compared to those inside, the information in the input data may not be sufficient to obtain a reasonable extrapolation. While if it is low, the number of new instances to be generated decreases, making the task too easy for our algorithm. After analyzing the results (Figure \ref{fig:crop_analysis}), we observed that the value of $k=0.5$ gives a good trade-off between the two extremes.

\begin{figure*}
\begin{center}
\includegraphics[width=0.9\linewidth]{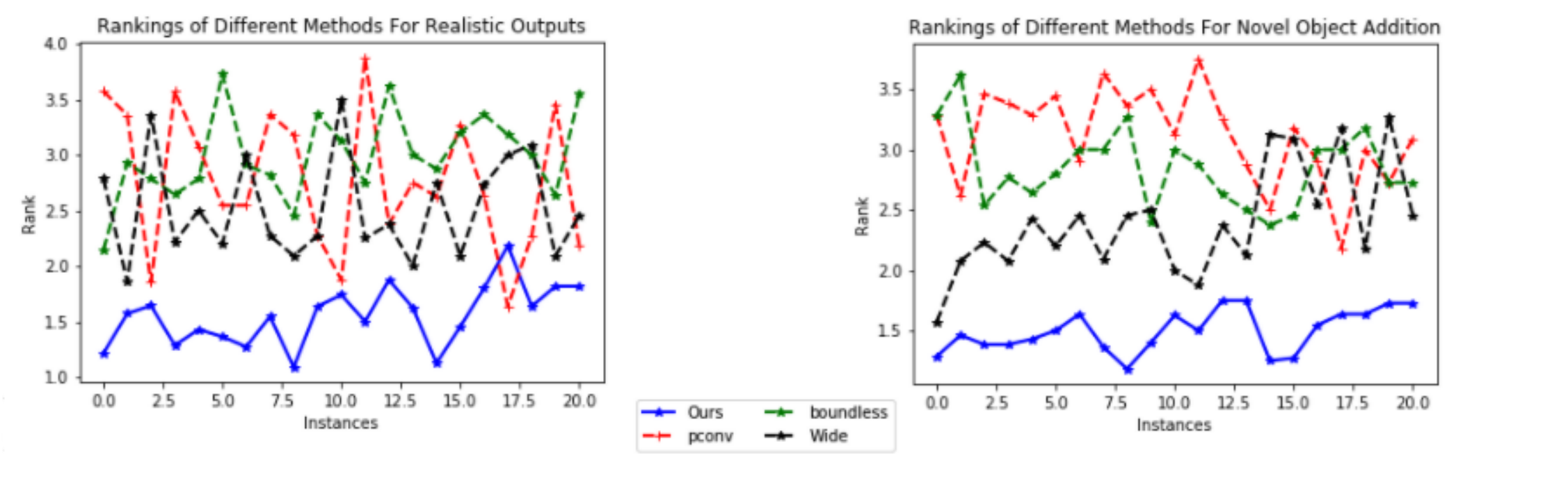}    
\end{center}
\caption{User preference study on the rank of various baselines and our method.
Lower rank means better.}
\label{fig:user-study_1}
\end{figure*}

\section{Human Evaluation}
We compare our method with the baselines also via human subjective study. The baselines included results from Partial Convolutions (PConv) \cite{liu2018image}, Boundless \cite{teterwak2019boundless} and OutPainting-SRN \cite{wang2019wide}. About hundred random people were given a set of 20 randomly selected images from both the datasets. They were given unlimited time to make the selection. They were asked to rank the images based on two parameters, viz.  Realistic appearance and  New object generation. Plot \ref{fig:user-study_1} shows the evaluation results. We found that our results were strongly favoured by most of the people. 

\begin{figure*}
\begin{center}
\includegraphics[width=0.9\linewidth]{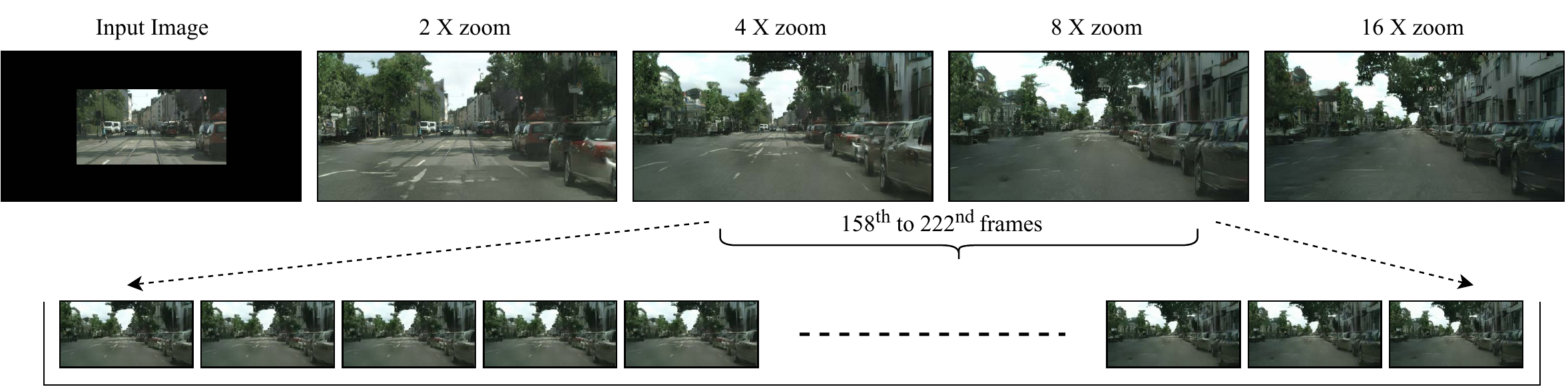} \end{center}
\caption{Given an input image, we extrapolate the image recursively to obtain $2\times, 4\times, 8\times, 16\times, 32\times, 64\times$ extrapolations. Using the extrapolated images, we obtain frames to make a video by performing naive `interpolation' between the extrapolated images. We achieve this by taking several crops of increasing sizes followed by bicubic downsampling to ensure all frames are of the same size. We show in the figure how frames are generated by `interpolating' between the $4\times$ and $8\times$ extrapolations.}
\label{fig:inf-zoom}
\end{figure*}

\section{Infinite Zoom}
\label{sec:inf_zoom}
Our model has ideally an infinite zooming out potential. We can zoom out an image upto a good extent by recursively passing through our model. We experimented this by training on modified Cityscapes dataset, made by removing the Mercedes at the bottom and then upsampling the images to the original dimension. We did this since if we ``infinitely" zoomed out the original images, then there would have been Mercedes recursively at the bottom in the zoomed out images.

As shown in Figure \ref{fig:inf-zoom}, the input image is first passed through our model to generate the output image, extrapolated 2 times in length and breadth. The output image thus generated is downsampled to half the size of the output image and is zero padded to generate the input image for the next iteration. We continue this process recursively 6 times, in course of which we generate the $2\times, 4\times \cdots$ till $64\times$ extrapolated images. To generate the frames for the video, after generating the $2\times$ extrapolated image, we downsample it gradually by 2 pixels in height and 4 pixels in width and zero pad it. Thus, for every input-output transition we generate 64 frames. There are 6 input-output transitions and hence, 384 frames. We also add 30 frames of the initial input at the beginning of the video for better realisation of the input image. The videos corresponding to some of the data images can be found in the project page \url{https://semie-iccv.github.io/}.

\section{Segmentation maps comparison}
\begin{figure*}
\begin{center}
\includegraphics[width=\linewidth]{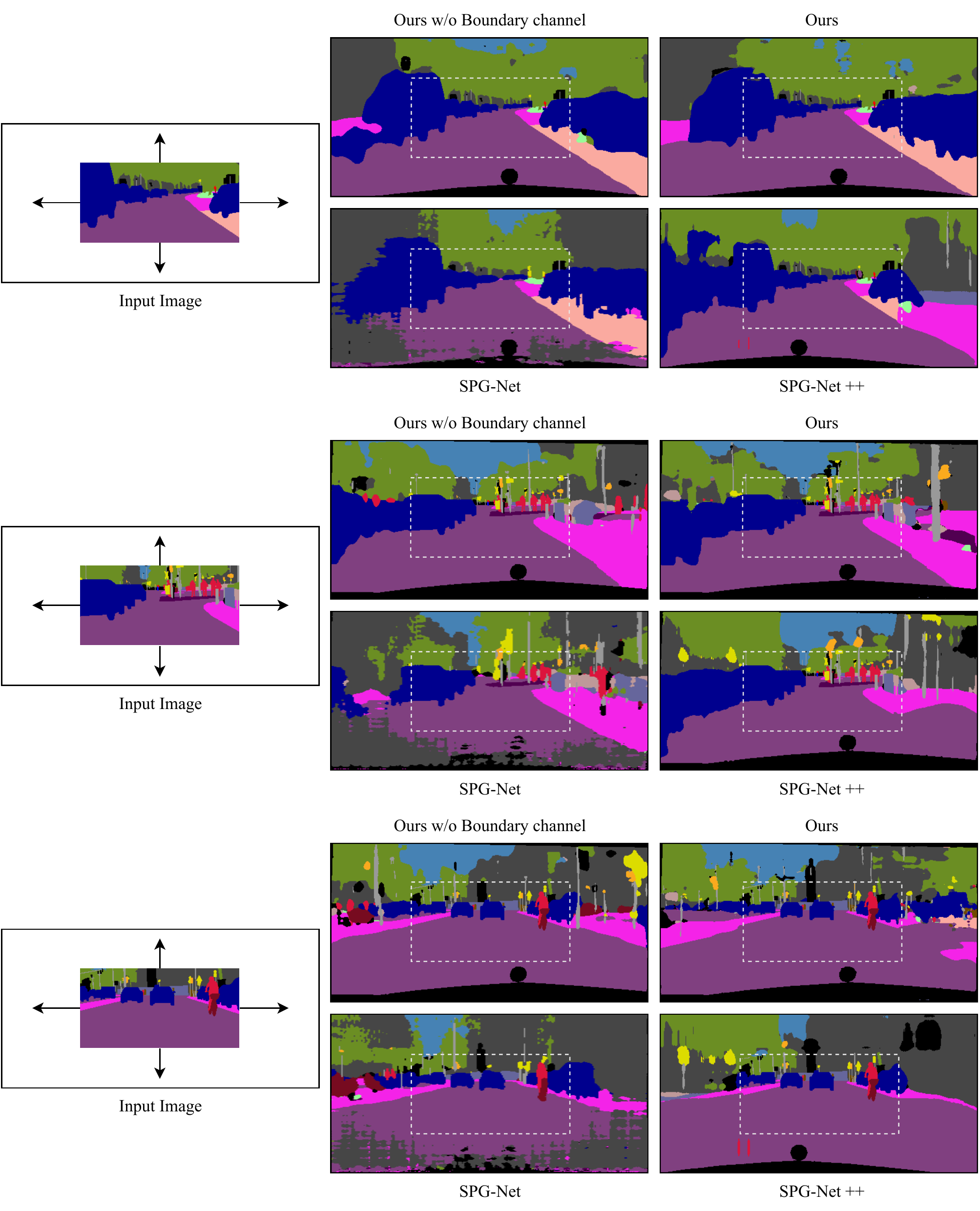}    
\end{center}
\caption{Additional semantic label extrapolation comparison. For each input image, the output of the PSPNet (stage 1) is shown in the leftmost column and the corresponding semantic extrapolation methods are shown. Top row shows results for the two variants of our method, with and without boundary channel regularizer. The bottom rows results for the two variants of the baseline, \cite{song2018spg} and SPG-Net++.}
\label{fig:semantic1}
\end{figure*}

\begin{figure*}
\begin{center}
\includegraphics[width=\linewidth]{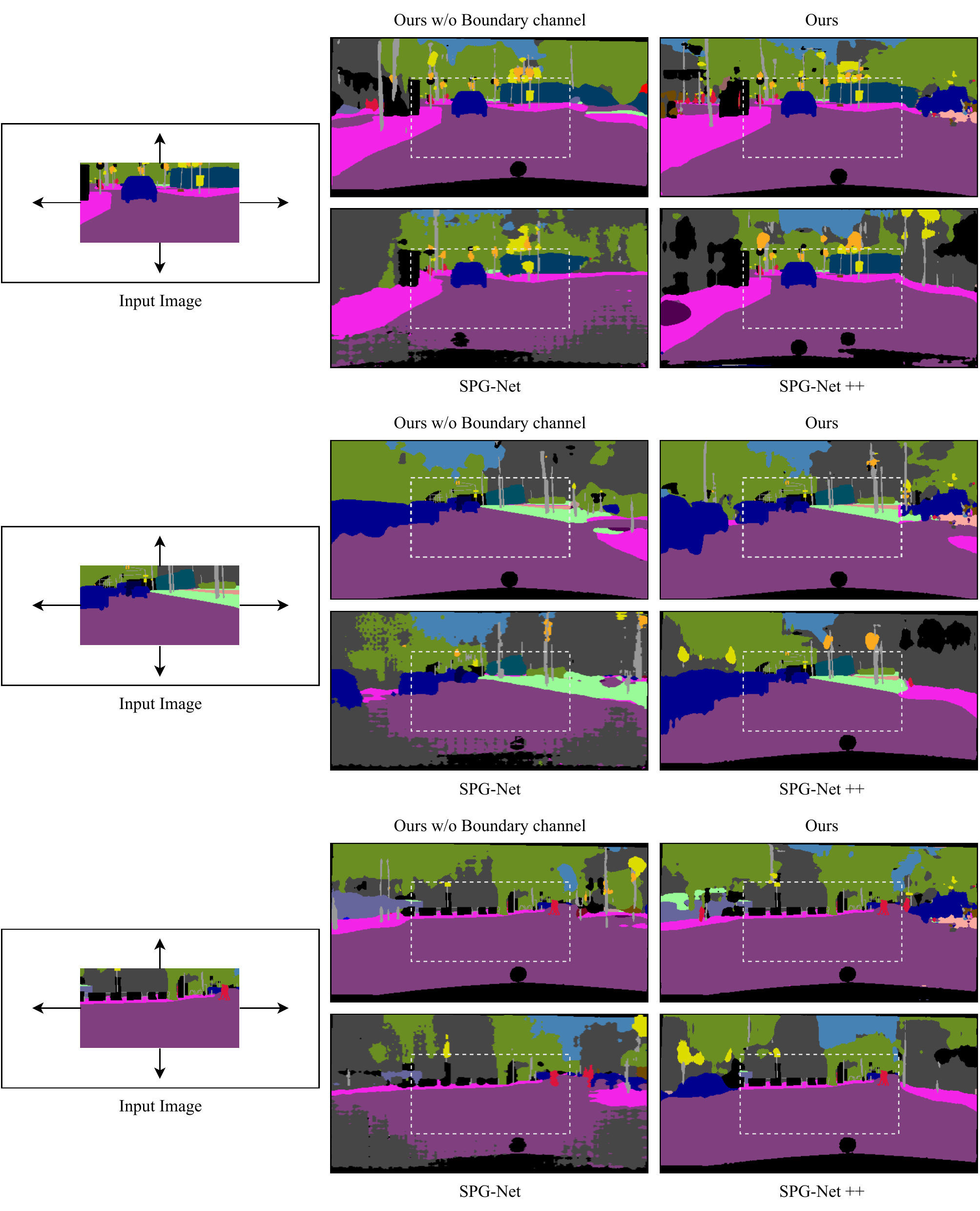}    
\end{center}
\caption{Additional semantic label extrapolation comparison. For each input image, the output of the PSPNet (stage 1) is shown in the leftmost column and the corresponding semantic extrapolation methods are shown. Top row shows results for the two variants of our method, with and without boundary channel regularizer. The bottom rows results for the two variants of the baseline, \cite{song2018spg} and SPG-Net++.}
\label{fig:semantic2}
\end{figure*}

We highlighted the importance of using extra boundary channel in semantic label extrapolation for better shapes of the instances. In Figure \ref{fig:semantic1} and \ref{fig:semantic2}, we compare our results with those generated without boundary channel. We further compare them with those generated using SPGNet \cite{song2018spg} and SPGNet++.

\section{Why do single stage approaches fail?}
\begin{figure*}
\begin{center}
\includegraphics[width=\linewidth]{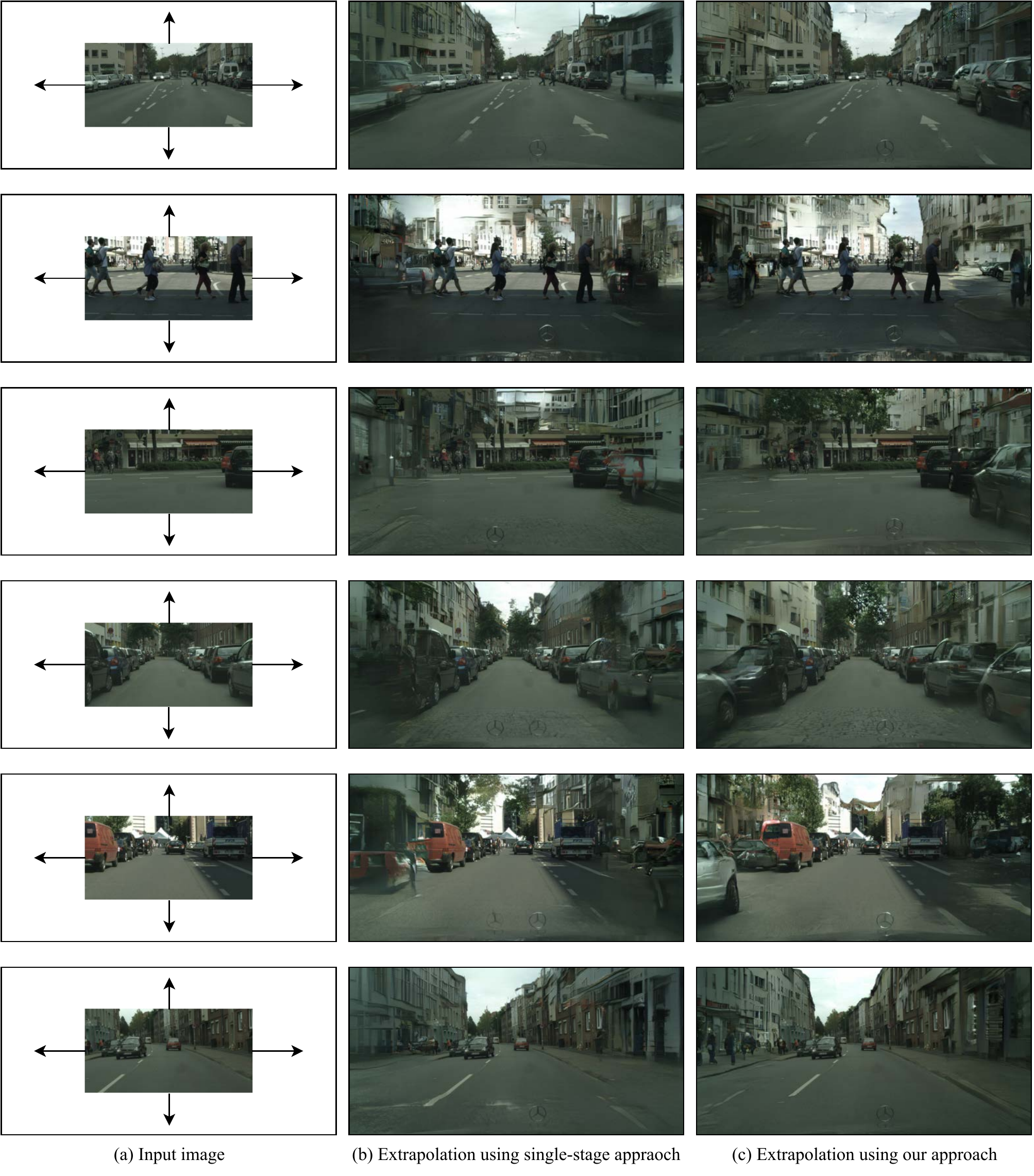}    
\end{center}
\caption{Comparison between ours and single stage approach: (a) Input image (b) Extrapolated image using our single stage approach (c) Extrapolated image using our current approach.}
\label{fig:im2im}
\end{figure*}

All the baseline methods except SPG-Net are based on direct image-to-image translation in a single stage. To analyse the strength of our approach and the failure of single stage approaches, we try direct image extrapolation by image-to-image translation. For this, during the training time, we used the RGB cropped input image and extrapolated it to the final RGB image using the stage 4, without IaCN and $D_{\text{patch}}$, directly (Figure \ref{fig:im2im}) using SPADE network. We observed that the images are generated with mere texture extension at the periphery with minimal to no new object generation.

\section{Additional results}
\begin{figure*}
\begin{center}
\includegraphics[width=\linewidth]{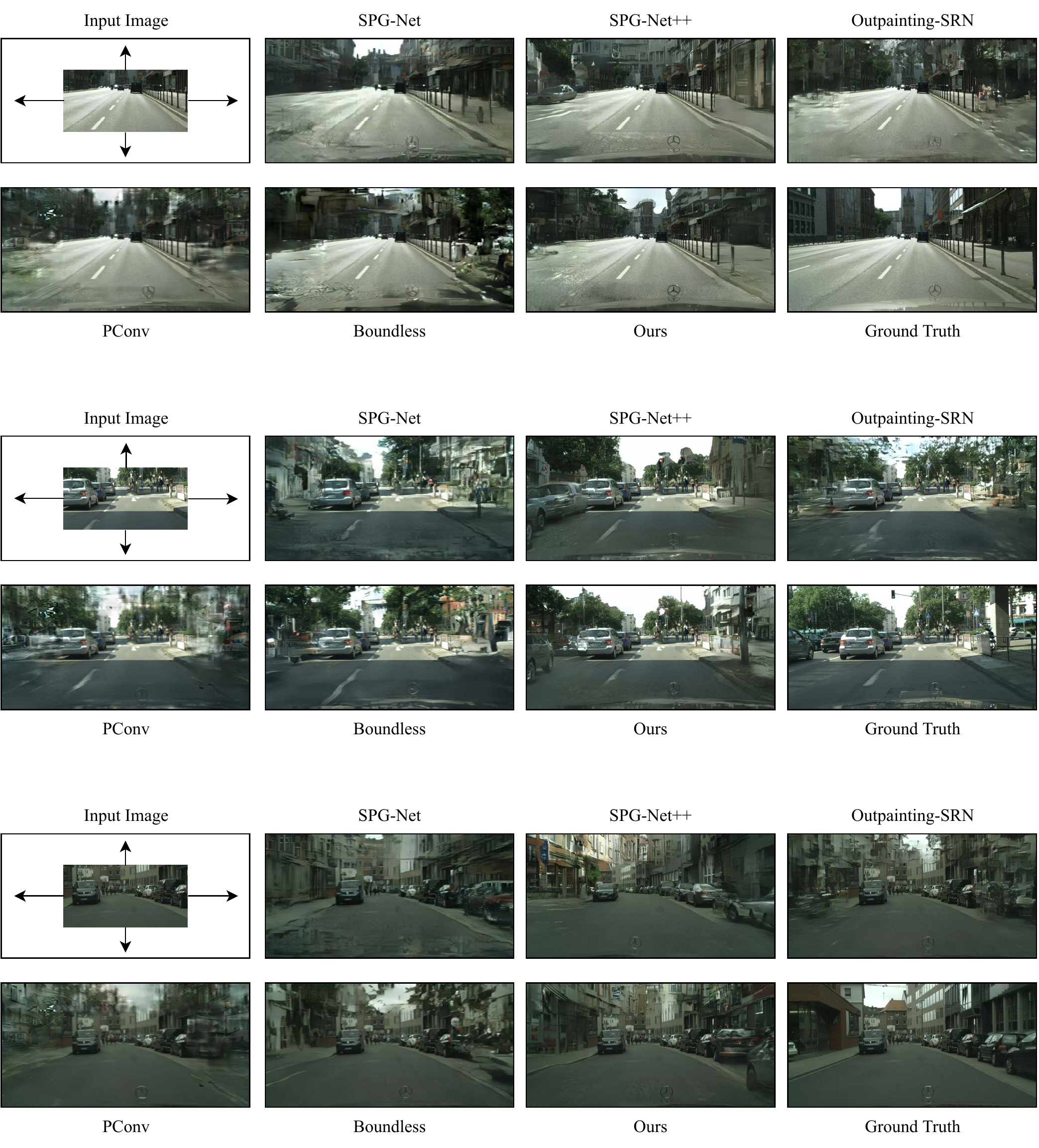}
\end{center}
  \caption{Additional comparisons between our model and the baselines on Cityscapes dataset. The baselines include results from Partial Convolutions (PConv) \cite{liu2018image}, Boundless \cite{teterwak2019boundless}, OutPainting-SRN \cite{wang2019wide}, SPGNet \cite{song2018spg}, and SPGNet++.}
\label{fig:city1}
\end{figure*}

\begin{figure*}
\begin{center}
\includegraphics[width=\linewidth]{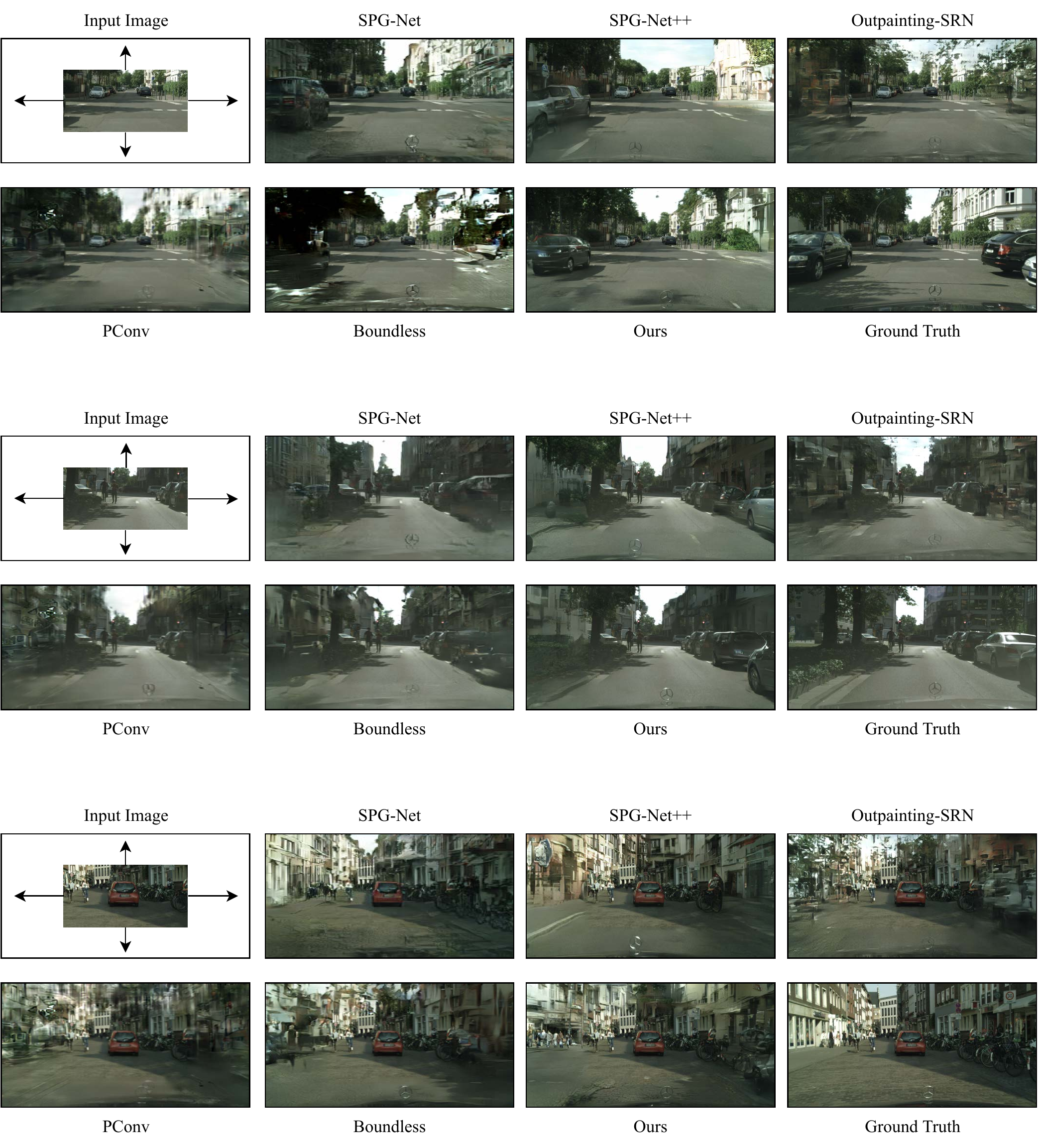}
\end{center}
  \caption{Additional comparisons between our model and the baselines on Cityscapes dataset. The baselines include results from Partial Convolutions (PConv) \cite{liu2018image}, Boundless \cite{teterwak2019boundless}, OutPainting-SRN \cite{wang2019wide} and SPGNet \cite{song2018spg}, SPGNet++.}
\label{fig:city2}
\end{figure*}

\begin{figure*}
\begin{center}
\includegraphics[width=0.9\linewidth]{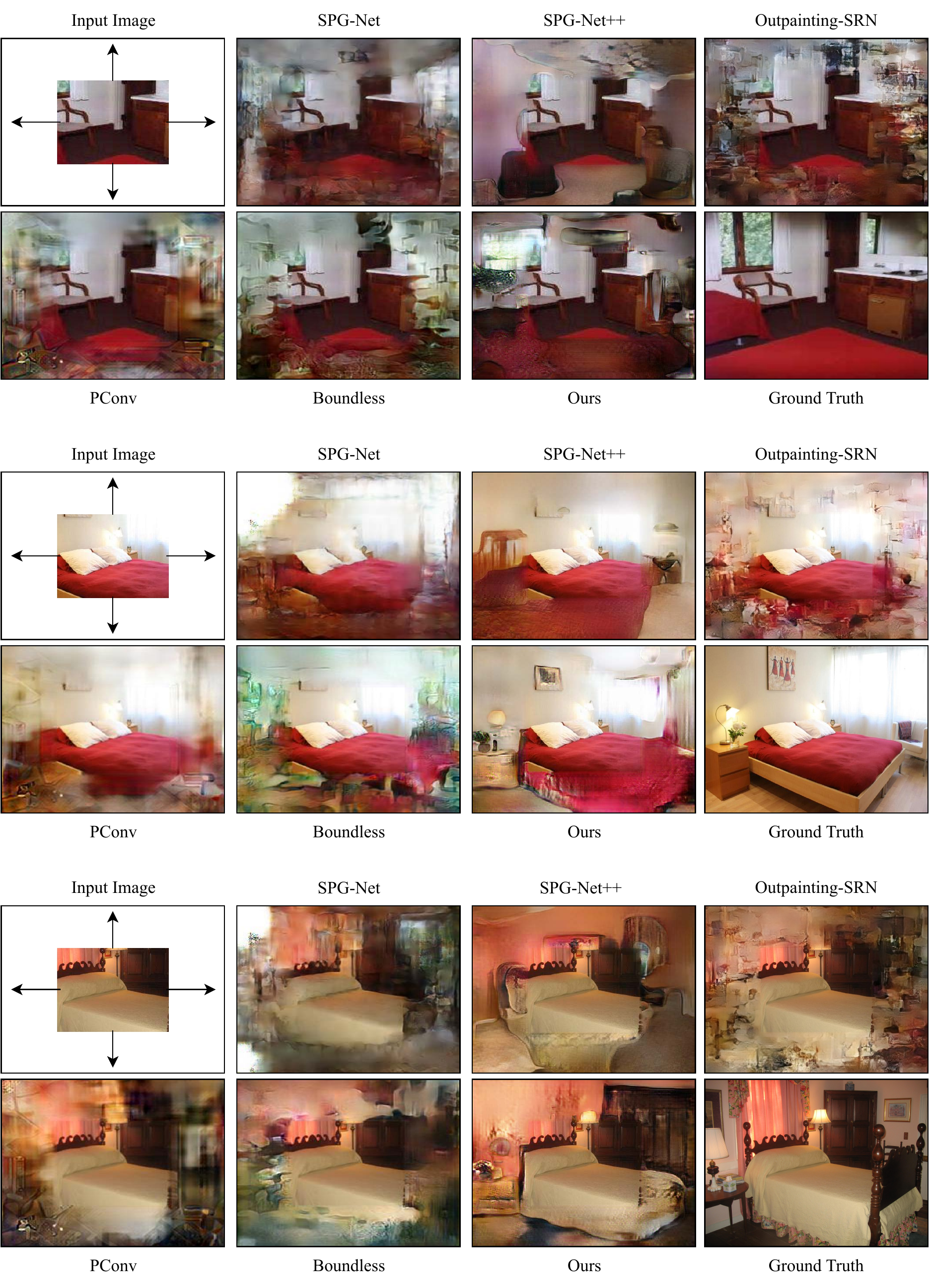}
\end{center}
  \caption{Additional comparisons between our model and the baselines on ADE-20K dataset. The baselines include results from Partial Convolutions (PConv) \cite{liu2018image}, Boundless \cite{teterwak2019boundless}, OutPainting-SRN \cite{wang2019wide}, SPGNet \cite{song2018spg}, and SPGNet++.}
\label{fig:ade1}
\end{figure*}

In Figures \ref{fig:city1}, \ref{fig:city2} and \ref{fig:ade1}, we show additional synthesized results and compare them with Partial Convolutions (PConv) \cite{liu2018image}, Boundless \cite{teterwak2019boundless}, OutPainting-SRN \cite{wang2019wide}, SPGNet \cite{song2018spg} and SPGNet++.

\end{document}